%% file: embedding.tex
\documentclass{article}

\usepackage{microtype}
\usepackage{graphicx}
\usepackage{subcaption}
\usepackage{booktabs} %
\usepackage{enumitem}
\usepackage{amssymb, amsmath}
\usepackage[labelfont=bf, textfont=bf]{caption}
\usepackage{multirow}
\usepackage{xspace}
\usepackage{microtype}
\usepackage{graphicx}
\usepackage{tabularx}
\usepackage{amssymb, amsmath}
\usepackage{multirow}
\usepackage{makecell}
\usepackage{xspace}
\usepackage[autostyle]{csquotes}
\usepackage{varwidth}
\usepackage{enumitem}
\usepackage{mwe}
\usepackage[export]{adjustbox}
\usepackage{balance}
\usepackage{fixltx2e}
\usepackage{amsthm}

\usepackage{url}

\usepackage{breakurl}
\usepackage[breaklinks]{hyperref}

\usepackage[firstpage]{draftwatermark}
\SetWatermarkText{
 \hspace*{2.5in}
 \raisebox{10in}{
  \includegraphics[height=0.9in]{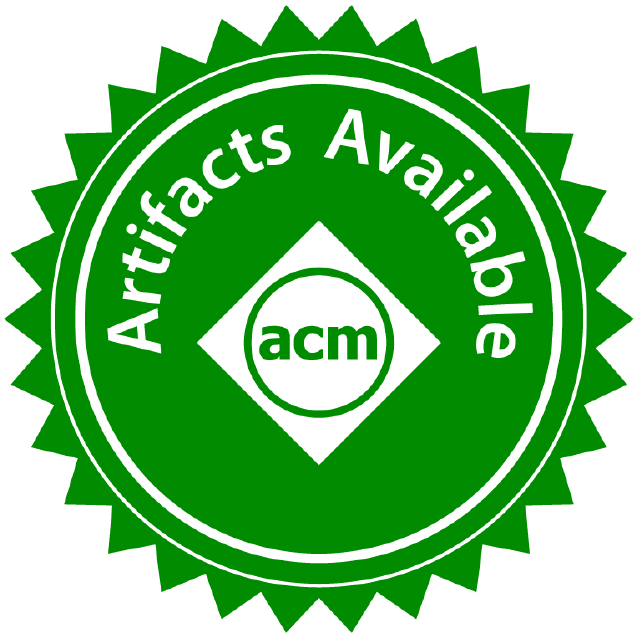}
  \includegraphics[height=0.9in]{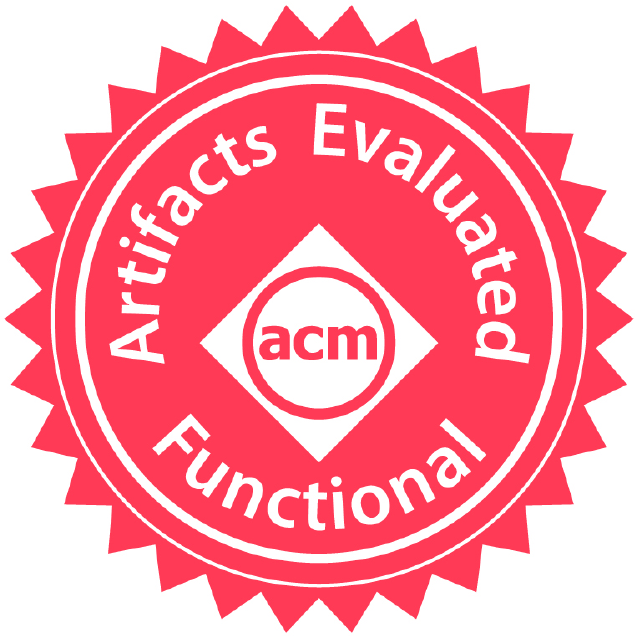}
  \includegraphics[height=0.9in]{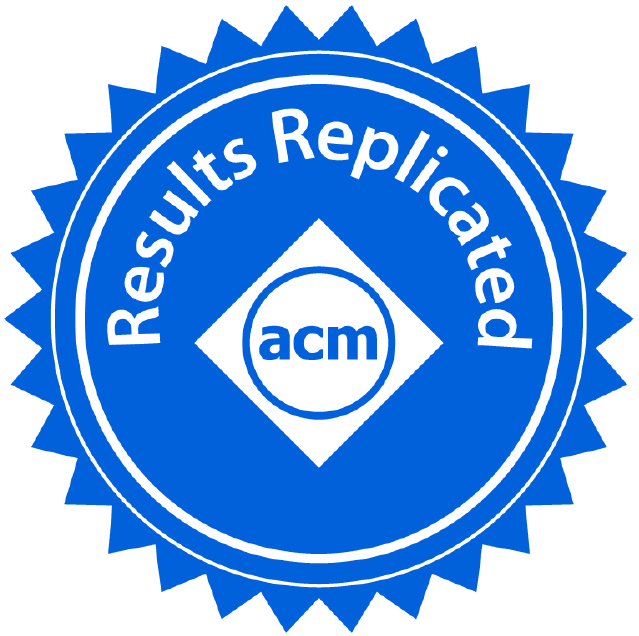}
 }
}
\SetWatermarkAngle{0}

\usepackage{hyperref}

\newtheorem{definition}{Definition}
\newtheorem{proposition}{Proposition}
\newtheorem{innercustomprop}{Proposition}
\newenvironment{customprop}[1]
{\renewcommand\theinnercustomprop{#1}\innercustomprop}
{\endinnercustomprop}

\newcommand{\tw}{\tilde{w}}
\newcommand{\tE}{\tilde{E}}
\newcommand{\tX}{\tilde{X}}
\newcommand{\tU}{\tilde{U}}
\newcommand{\tS}{\tilde{S}}
\newcommand{\tR}{\tilde{R}}
\newcommand{\tV}{\tilde{V}}
\newcommand{\tW}{\tilde{W}}

\newcommand{\tf}{\tilde{f}}
\newcommand{\tx}{\tilde{x}}

\newcommand{\RR}{\mathbb{R}}
\providecommand{\tr}{\mathop{\rm tr}}
\newcommand{\ProbOpr}[1]{\mathbb{#1}}
\newcommand{\expect}[2]{%
	\ifthenelse{\equal{#2}{}}{\ProbOpr{E}_{#1}}
	{\ifthenelse{\equal{#1}{}}{\ProbOpr{E}\left[#2\right]}{\ProbOpr{E}_{#1}\left[#2\right]}}} %
\newcommand{\defeq}{:=}
\newcommand{\eqdef}{=:}

\newcommand{\yell}[1]{{\color{black}#1}}

\newcommand{\ie}{i.e.}

\newcolumntype{Y}{>{\centering\arraybackslash}X}

\newcommand{\m}{{matrix completion}\xspace}

\usepackage[accepted]{mlsys2020}

\mlsystitlerunning{Understanding the Downstream Instability of Word Embeddings}

\begin{document}

\twocolumn[
\mlsystitle{Understanding the Downstream Instability of Word Embeddings}

\mlsyssetsymbol{equal}{*}

\begin{mlsysauthorlist}
\mlsysauthor{Megan Leszczynski}{st}
\mlsysauthor{Avner May}{st}
\mlsysauthor{Jian Zhang}{sn}
\mlsysauthor{Sen Wu}{st}
\mlsysauthor{Christopher R. Aberger}{sn}
\mlsysauthor{Christopher R{\'e}}{st}
\end{mlsysauthorlist}

\mlsysaffiliation{st}{Department of Computer Science, Stanford University}
\mlsysaffiliation{sn}{SambaNova Systems}

\mlsyscorrespondingauthor{Megan Leszczynski}{mleszczy@stanford.edu}

\mlsyskeywords{Machine Learning, MLSys}

\vskip 0.3in

\input{abstract}
]

\printAffiliationsAndNotice{{}} %

\input{introduction}
\input{preliminaries}
\input{characterizing}
\input{measuring}
\input{evaluation}
\input{extensions}
\input{related}

\input{conclusion}

\section*{Acknowledgements}

We thank Charles Kuang, Shoumik Palkar, Fred Sala, Paroma Varma, and the anonymous reviewers for their valuable feedback.
We gratefully acknowledge the support of DARPA under Nos. FA87501720095 (D3M), FA86501827865 (SDH), and FA86501827882 (ASED); NIH under No. U54EB020405 (Mobilize), NSF under Nos. CCF1763315 (Beyond Sparsity), CCF1563078 (Volume to Velocity), and 1937301 (RTML);  ONR under No. N000141712266 (Unifying Weak Supervision); the Moore Foundation, NXP, Xilinx, LETI-CEA, Intel, IBM, Microsoft, NEC, Toshiba, TSMC, ARM, Hitachi, BASF, Accenture, Ericsson, Qualcomm, Analog Devices, the Okawa Foundation, American Family Insurance, Google Cloud, Swiss Re, NSF Graduate Research Fellowship under No. DGE-1656518, and members of the Stanford DAWN project: Teradata, Facebook, Google, Ant Financial, NEC, VMWare, and Infosys. The U.S. Government is authorized to reproduce and distribute reprints for Governmental purposes notwithstanding any copyright notation thereon. Any opinions, findings, and conclusions or recommendations expressed in this material are those of the authors and do not necessarily reflect the views, policies, or endorsements, either expressed or implied, of DARPA, NIH, ONR, or the U.S. Government.

\bibliographystyle{mlsys2020}
\bibliography{embedding}

\appendix
\onecolumn

\input{artifact}

\input{theory}
\input{extended_setup}
\input{extended_empirical}
\input{limitations}

\end{document}

%% file: abstract.tex
\begin{abstract}

Many industrial machine learning (ML) systems require frequent retraining to keep up-to-date with constantly changing data.
This retraining exacerbates a large challenge facing ML systems today: model training is unstable, i.e., small changes in training data can cause significant changes in the model's predictions.
In this paper, we work on developing a deeper understanding of this instability, with a focus on how a core building block of modern natural language processing (NLP) pipelines---pre-trained word embeddings---affects the instability of downstream NLP models. We first empirically reveal a tradeoff between stability and memory: increasing the embedding memory \yell{$2\times$} can reduce the disagreement in predictions due to small changes in training data by \yell{5\% to 37\%} (relative).
To theoretically explain this tradeoff, we introduce a new measure of embedding instability---the eigenspace instability measure---which we prove bounds the disagreement in downstream predictions introduced by the change in word embeddings.
Practically, we show that the eigenspace instability measure can be a cost-effective way to choose embedding parameters to minimize instability without training downstream models, outperforming other embedding distance measures and performing competitively with a nearest neighbor-based measure.
Finally, we demonstrate that the observed stability-memory tradeoffs extend to other types of embeddings as well, including knowledge graph and contextual word embeddings.

\end{abstract}

%% file: introduction.tex
\section{Introduction}

Data is more dynamic than ever before: every input, interaction, and response is captured and archived in hopes of extracting insights with machine learning (ML) models. To stay up-to-date, models must be frequently retrained, with the freshness of models becoming a requirement for user satisfaction in numerous products, from ads~\cite{He2014PracticalLF} to recommendation systems~\cite{Covington2016DeepNN}.
However, frequent retraining can lead to large and unwanted fluctuations in model predictions due to the \textit{instability} of many machine learning training algorithms: minimal changes in training data can produce significantly different predictions~\cite{Fard2016LaunchAI}.
From discussions with engineers in an e-commerce firm, an online social media company, and a Fortune 500 software company, we found that instability from retraining is one of their largest, and also most under-addressed, pain points.
As a result of instability, ML engineers struggle to identify genuine concept shifts, spend more time tracking down regressions, and require more resources retraining downstream model dependencies. Diagnosing and reducing instability in a cost-effective way is a major challenge for today's machine learning pipelines.

In this work, we take a first step toward addressing the problem of ML model instability by examining in detail a core building block of most modern natural language processing (NLP) applications: word embeddings~\cite{mikolov2013a,mikolov2013b,pennington2014glove,bojanowski2016enriching}.
Several recent works have shown that word embeddings are unstable, with the nearest neighbors to words varying significantly across embeddings trained under different settings~\cite{hellrich2016bad,antoniak2018evaluating,wendlandt18surprising,Pierrejean2018PredictingWE,chugh,Hellrich2019TheIO}.
These results may cause researchers using embeddings for analysis to reassess the reliability of their conclusions.
Moreover, these results raise questions about \emph{how the embedding instability impacts downstream NLP tasks}---an area which remains largely unexplored and which we focus on in this work.
We define the downstream instability between a pair of word embeddings as the percentage of predictions which change between the models trained on the two embeddings for a given task.
By this notion of instability, we find that \yell{15\%} of predictions on a sentiment analysis task can disagree due to training the embeddings on an accumulated dataset with just 1\% more data.
In embedding servers, where an embedding is reused among multiple downstream tasks~\cite{uberblog,tfhub,twitterblog,googleblog}, the impact of this instability can be quickly amplified.
Understanding this downstream instability is challenging, however, because it requires \emph{both theoretical and empirical} insights on how the embedding instability propagates to the downstream tasks.

The goal of this paper is to develop a deeper understanding of the downstream instability of word embeddings.
This understanding could both drive the design choices for embedding systems (i.e. choosing hyperparameters) and lead to efficient techniques to distinguish among unstable and stable embeddings without training downstream models.
To achieve this, we perform a study on the downstream instability of word embeddings across multiple embedding algorithms and downstream tasks.
Our study exposes a novel trade-off between stability and another critical property of embeddings---\emph{memory}.
We find that increasing the memory can lead to more stable embeddings, with a \yell{2$\times$} increase in memory reducing the percentage prediction disagreement on downstream tasks by \yell{5\% to 37\%} (relative).
Determining how the memory affects the instability is not straightforward: factors like the \emph{dimension}, a hyperparameter controlling the expressiveness of the embedding, and the \emph{precision}, the number of bits used per entry in the embedding after compression, can independently affect the instability and interact in unexpected ways.
To better understand the stability-memory tradeoff empirically, we study the effects of dimension and precision both in isolation and together.
This important stability-memory tradeoff leads us to ask two key questions: (1) theoretically, how can we explain this tradeoff, and (2) practically, how can we select the dimension-precision\footnote{For brevity, we refer to a pair of dimension and precision parameters as the ``dimension-precision" parameters.} parameters to minimize the downstream instability?

To theoretically explain the stability-memory trade-off, we introduce a new measure for embedding instability---the eigenspace instability measure---which we theoretically relate to downstream instability in the case of linear regression models.  The eigenspace instability measure builds on the eigenspace overlap score~\cite{smallfry}, and measures the degree of similarity between the eigenvectors of the Gram matrices of a pair of embeddings, weighted by their eigenvalues. We show that the expected downstream disagreement between the linear regression models trained on two embedding matrices can be expressed in terms of the eigenspace instability measure. Furthermore, these theoretical insights have a practical application: we propose using the eigenspace instability measure to efficiently select dimension-precision parameters with low downstream instability, without having to train downstream models.

We empirically validate that the eigenspace instability measure correlates strongly with the downstream instability and that the measure is effective as a selection criterion for the dimension-precision parameters.
First, we show that the theoretically grounded eigenspace instability measure more strongly correlates with downstream instability than the majority of the other embedding distance measures (i.e. semantic displacement~\cite{hamilton2016}, the PIP loss~\cite{Yin2018OnTD}, and the eigenspace overlap score~\cite{smallfry}) and attains Spearman correlations from \yell{0.04} better to \yell{0.09} worse than the other top-performing measure, the k-NN measure (e.g., ~\citet{hellrich2016bad,antoniak2018evaluating,wendlandt18surprising}), which lacks theoretical guarantees.
Next, we show that when using an embedding distance measure to choose the more stable dimension-precision parameters out of a pair of choices, the eigenspace instability measure achieves up to \yell{3.33$\times$} lower error rates than the weaker measures and from \yell{$0.95\times$ to $1.55\times$} the error rate of the k-NN measure.
On the more challenging task of selecting the combination of dimension and precision under a  memory budget, we show that eigenspace instability measure attains a difference in prediction disagreement to the oracle up to \yell{2.98\%} (absolute) better than the weaker baselines and within \yell{0.35\%} (absolute) of the k-NN measure.

To summarize, we make the following contributions:
\vspace{-1em}
\begin{itemize}[itemsep=0mm]
	\item We study the downstream instability of word embeddings, revealing a novel stability-memory tradeoff. In particular, we study the impact of two key parameters, dimension and precision, and propose a simple rule of thumb relating the embedding memory and downstream instability (Section~\ref{sec:tradeoffs}).
	\item To theoretically explain this tradeoff, we introduce a new measure for embedding instability, the eigenspace instability measure, that we prove theoretically determines the expected downstream disagreement on a linear regression task (Section~\ref{sec:measuring}).
	\item To empirically validate our theory, we perform an evaluation of methods for selecting embedding hyperparameters to minimize downstream instability.
	Practically, we show that the eigenspace instability measure can outperform the majority of other embedding distance measures and perform similarly to the k-NN measure, for which we have no theoretical guarantees.
	\item  Finally, we show that the stability-memory tradeoffs extend to knowledge graph embeddings~\cite{bordes2013translating} and contextual word embeddings, such as BERT embeddings~\cite{Devlin2018BERTPO}. For instance, we find that increasing the memory of knowledge graph embeddings 2$\times$ decreases the instability on a link prediction task by \yell{7\% to 19\%} (relative) (Section~\ref{sec:extensions}).
\end{itemize}

%% file: preliminaries.tex
\section{Preliminaries}
\label{sec:prelim}
We begin by formally defining the notion of instability we use in this work. We then review the word embedding algorithms and compression technique used in our study, and discuss existing measures to compare two embeddings.

\subsection{Instability Definition}
\label{def}
We define the downstream instability as follows:

\begin{definition}
Let $X \in \RR^{n \times d}$ and $\tilde{X} \in \RR^{n \times k}$ be two embedding matrices, and let $f_X$ and $f_{\tilde{X}}$ represent models trained using X and $\tilde{X}$, respectively, for a downstream task T.  Then the instability between X and $\tilde{X}$ with respect to task T is defined as
\vspace{-3mm}
\begin{eqnarray*}
    \mathcal{DI}_T(X,\tilde{X}) = \frac{1}{N} \sum_{i=1}^N \mathcal{L}(f_X(z_i), f_{\tilde{X}}(z_i)),
\end{eqnarray*}
\vspace{-3mm}

 where $\{z_i\}_{i=1}^N$ is a heldout set for task T, and $\mathcal{L}$ is a fixed loss function.
\end{definition}
When the zero-one loss is used for $\mathcal{L}$, this measure captures the percentage of predictions which disagree on downstream models trained on each embedding.

\subsection{Word Embedding Algorithms}
\label{wordembeddings}

Word embedding algorithms learn distributed representations of words by taking as input a textual corpus $C$ and returning the word embedding $X \in \mathbb{R}^{n \times d}$, where $d$ is the dimension of the embeddings and $n$ is the vocabulary size.
We evaluate \m (MC)~\cite{jin2016provable}, GloVe~\cite{pennington2014glove}, and continuous bag-of-words (CBOW)~\cite{mikolov2013a,mikolov2013b} embedding algorithms.
MC and GloVe factor the co-occurrence matrix $A \in \mathbb{R}^{n \times n}$ generated from $C$, whereas CBOW operates on the sequential corpus $C$ directly. We elaborate below.

\paragraph*{Matrix completion (MC)} Matrix completion uses the word embeddings to approximate the observed word co-occurrence $A$ and can be formally written as:
\begin{equation*}
\begin{aligned}
& V = \arg\min_{X} \sum_{(i, j) \in \Theta} (X_i X_j^T - A_{ij})^2
\end{aligned}
\label{sgdeqn}
\end{equation*}
\noindent where $\Theta$ are the observed (non-zero) entries in $A$. Following standard technique, $A$ is the positive pointwise mutual information (PPMI) matrix, rather than the true co-occurrence matrix~\cite{Bullinaria2007ExtractingSR}.

We solve the matrix completion problem using an online algorithm similar to that proposed in \citet{jin2016provable}. We iteratively train $X$ via stochastic gradient descent (SGD) after computing the loss on sampled entries of the observed co-occurrence matrix $A$.

\paragraph*{GloVe} Similar to MC, GloVe solves a matrix factorization problem, but approximates the co-occurrence information in a weighted form to reduce noise from rare co-occurrences. GloVe models the word and context embeddings separately.

\paragraph*{Continuous bag-of-words (CBOW)} The CBOW algorithm predicts a word given its local context words. The embedding matrix $X$ is trained via SGD, where the loss maximizes the probability that an observed word and context pair co-occurs in the corpus and minimizes the probability that a negative sample co-occurs.
We use the \texttt{word2vec} implementation of CBOW.\footnote{\url{https://github.com/tmikolov/word2vec}}

\vspace{-2mm}

\subsection{Compression Technique}

We use a standard technique---uniform quantization---to compress word embeddings. Recent work~\cite{smallfry} demonstrates that uniform quantization performs on par in terms of downstream quality with more complex compression techniques, such as k-means compression~\cite{Andrews2015CompressingWE} and deep compositional code learning~\cite{shu2018compressing}. We leverage their implementation\footnote{\url{https://github.com/HazyResearch/smallfry}} to apply uniform quantization to word embeddings to study the impact of the precision on instability. Under uniform quantization, each entry in the word embedding matrix is rounded to a discrete value in a set of $2^b$ equally spaced values within an interval,  such that each entry can be represented with just $b$ bits.
For more details on the way we use uniform quantization for our experiments, see Appendix~\ref{sec:compression-details}.

\subsection{Embedding Distance Measures}
\label{sec:baselines}
We consider four embedding distance measures from the literature to quantify the differences between embeddings. For each measure, we assume we have a pair of embeddings $X \in\mathbb R^{n\times d}$ and $\tilde X \in\mathbb R^{n\times d}$ trained on corpora $C$ and $\tilde C$, respectively, where $n$ is the size of the vocabulary and $d$ is the dimension of the embedding. Due to computational efficiency and our observation that downstream tasks use a majority of high frequency words, we only consider the top 10k most frequent words to compute each measure (including the eigenspace instability measure).

\vspace{-2mm}
\paragraph*{k-Nearest Neighbors (k-NN) Measure} Variants of the k-NN measure were used in recent works on word embedding stability to characterize the intrinsic stability of embeddings (e.g., ~\citet{hellrich2016bad,antoniak2018evaluating,wendlandt18surprising}).
The k-NN measure is defined as $\frac{1}{Q}\sum_{q=0}^{Q}\frac{| \textrm{N}_{k}(X; q) \cap{} \textrm{N}_{k}(\tilde{X}; q)|}{k}$, where $Q$ is the number of randomly sampled query words (we use $Q$=1000), and the $\textrm{N}_{k}$ function takes an embedding and the index of a query word, and returns the indices of the $k$ most similar words to the query word by the cosine distance.

\vspace{-2mm}
\paragraph*{Semantic Displacement} Researchers have used semantic displacement to compute the distance that words have shifted over time~\cite{hamilton2016}.
Semantic displacement can be defined as $\frac{1}{n}\sum_{i=0}^{n}\textrm{cos-dist}(X_i, R\tilde{X_i})$, where $R = \arg\min_\Omega ||X - \tilde{X}\Omega||_F$, subject to $\Omega^T\Omega=I$ (i.e., the orthogonal Procrustes solution~\cite{schonemann1966}).

\vspace{-2mm}
\paragraph*{Pairwise Inner Product Loss} The Pairwise Inner Product (PIP) loss was proposed for dimensionality selection to optimize for the intrinsic quality of an embedding~\cite{Yin2018OnTD}. The PIP loss is defined as $\|XX^T - \tilde{X}\tilde{X}^T \|_F$.

\paragraph*{Eigenspace Overlap Score} The eigenspace overlap score was recently proposed as a measure of compression quality~\cite{smallfry}. The eigenspace overlap is defined as $\frac{1}{d}\|U^T\tilde{U}\|^2_F$, where $X=USV^T$ and $\tilde{X} = \tilde{U}\tilde{S}\tilde{V}^T$ are the singular value decompositions (SVDs) of $X$ and $\tilde{X}$.

%% file: characterizing.tex
\section{A Stability-Memory Tradeoff}
\label{sec:tradeoffs}

We now present the empirical study that exposes the tradeoff we observe between downstream stability and embedding memory, and demonstrate that as the memory increases, the instability decreases.
We consider the dimension and precision of the embedding as two important axes controlling the memory of the embedding.
We first study the impact of the embedding's dimension and precision on downstream instability in isolation in Sections~\ref{sec:knob_dim} and \ref{sec:knob_comp}, respectively, followed by a discussion of their joint effect in Section~\ref{sec:joint}.
\vspace{-1mm}
\paragraph*{Corpora}

We use two full Wikipedia dumps\footnote{\url{https://dumps.wikimedia.org}}: Wiki'17 and Wiki'18, which we collected approximately a year apart, to train embeddings.
The corpora are pre-processed by a Facebook script\footnote{\url{https://github.com/facebookresearch/fastText/blob/master/get-wikimedia.sh}}, which we modify to keep the letter cases.
We use these two corpora as examples of the temporal changes which can occur to the text corpora used to train word embeddings.
Each corpora has about \yell{4.5} billion tokens, and when training the embeddings, we only learn the embeddings for the top 400k most frequent words.

\paragraph*{Downstream NLP Tasks}
After training the word embeddings, we compress the embeddings with uniform quantization and train models for downstream NLP tasks on top of the embeddings, fixing the embeddings during training. We train word embeddings with three seeds, and use the same corresponding seeds for the downstream models. Results are reported as averages over the three seeds, with error bars indicating the standard deviation. We also align all pairs of Wiki'17 and Wiki'18 embeddings (same dimension and seed) with orthogonal Procrustes~\cite{schonemann1966} prior to compressing and training downstream models, as preliminary experiments found this helped to decrease instability. For each downstream task, we perform a hyperparameter search for the learning rate using 400-dimensional Wiki'17 embeddings, and use the same learning rate across all dimensions to minimize the impact of learning rate on our analysis. Here, we discuss the two standard downstream NLP tasks we consider throughout our paper. Please see Appendix~\ref{sec:ds-setup} for more experimental setup details.

\textit{Sentiment Analysis.} We evaluate on a binary sentiment analysis task where given a sentence, the model determines if the sentence is positive or negative~\cite{Kim2014ConvolutionalNN}. We train a linear bag-of-words model for this task and evaluate on four benchmark datasets: MR~\cite{panglee}, MPQA~\cite{Wiebe2005AnnotatingEO}, Subj~\cite{Pang2004ASE}, and SST-2~\cite{Socher2013RecursiveDM}. We will be showing results on SST-2; for more results, see Appendix~\ref{sec:ext-trends}.

\textit{Named Entity Recognition (NER).} The named entity recognition task is a multi-class classification task to predict whether each token in the dataset is an entity, and if so, what type. We use a BiLSTM model~\cite{Akbik2018ContextualSE} for this task and evaluate on the benchmark CoNLL-2003 dataset~\cite{tjongkimsang2003conll}. Each token is assigned an entity label of PER, ORG, LOC, and MISC, or an O label, indicating outside of any entities (i.e., no entity). We measure instability only over the tokens for which the true value is an entity.  We use the BiLSTM without the conditional random field (CRF) decoding layer for computational efficiency; in Appendix~\ref{sec:complex-ds} we show that the trends also hold on a subset of the results with a BiLSTM-CRF.

\begin{figure}[t]
    \centering
    \includegraphics[width=0.95\columnwidth]{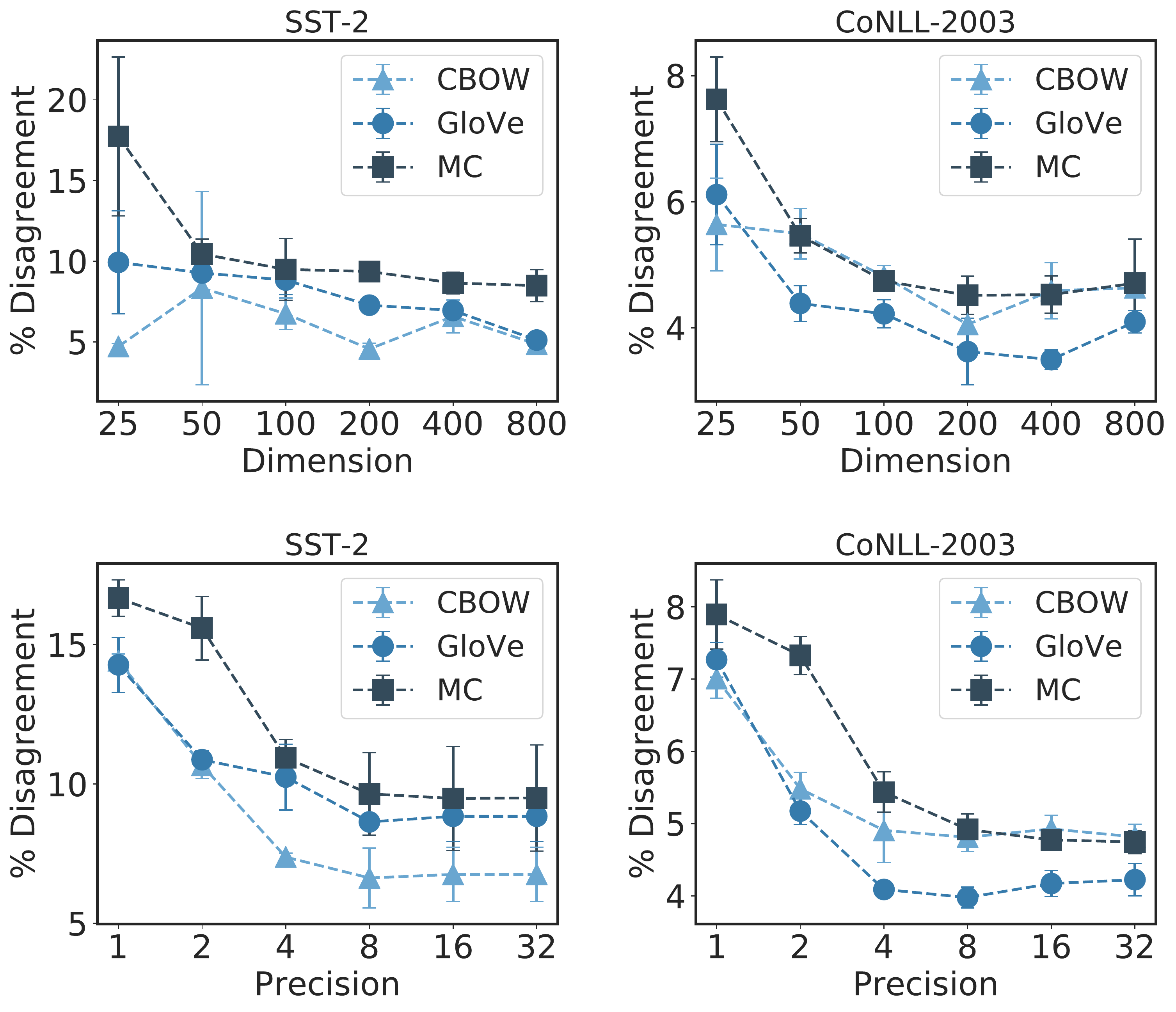}
    \caption{Downstream instability of sentiment analysis (SST-2) and NER (CoNLL-2003) tasks under different dimensions (top) and precisions (bottom) for CBOW, GloVe, and MC embeddings.}
    \vspace{-1.1em}
    \label{fig:dim-compress}
\end{figure}

\begin{figure*}[t]
    \centering
    \includegraphics[width=0.9\textwidth]{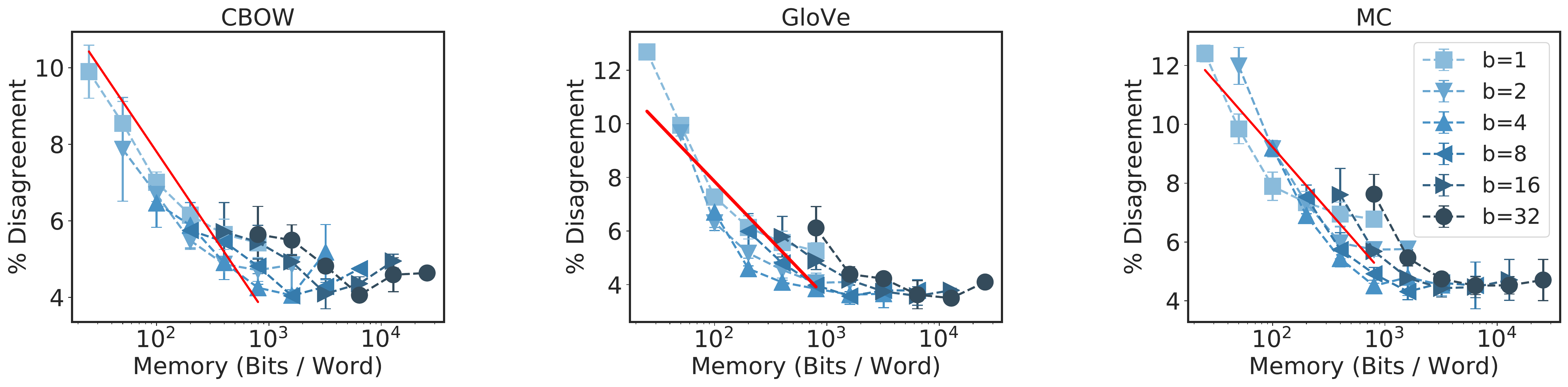}
    \caption{Downstream instability of NER (CoNLL-2003) tasks for various memory budgets with different dimension-precision combinations. The red line indicates the average linear-log model relating instability and memory.}
    \vspace{-1mm}
    \label{fig:joint}
    \end{figure*}

\subsection{Effect of Dimension}
\label{sec:knob_dim}

\label{sec:dim}
We evaluate the impact of the dimension of the embedding on its downstream stability, and show that generally as the dimension increases, the instability decreases.

\paragraph*{Tradeoffs}
To perform our tradeoff study, we train Wiki'17 and Wiki'18 embeddings with dimensions in \{25, 50, 100, 200, 400, 800\}, and train downstream models on top of the embeddings. We compute the prediction disagreement between models trained on Wiki'17 and Wiki'18 embeddings of the same dimension. In Figure~\ref{fig:dim-compress} (top), we see that as the dimension increases, the downstream instability often decreases across embedding algorithms and downstream tasks, plateauing at larger dimensions. In Section~\ref{sec:joint}, we see that these trends are even more pronounced in lower memory regimes when we also consider different precisions.

\vspace{-1mm}
\subsection{Effect of Precision}
\label{sec:knob_comp}

\label{sec:compression}
We evaluate the effect of the precision, the number of bits used to store each entry of the embedding matrix, on the downstream stability, and show that as the precision increases, the instability decreases.

\vspace{-1.5mm}
\paragraph*{Tradeoffs}
We compress 100-dimensional Wiki'17 and Wiki'18 embeddings with uniform quantization to precisions $b \in \{1, 2, 4, 8, 16, 32\}$,\footnote{$b=32$ signifies full-precision embeddings.} and train downstream models on top of the compressed embeddings. We compute the prediction disagreement between models trained on Wiki'17 and Wiki'18 embeddings of the same precision. In Figure~\ref{fig:dim-compress} (bottom), we show that as the precision increases, the instability generally decreases on sentiment analysis and NER tasks for CBOW, GloVe, and MC embedding algorithms. Moreover, we see that for precisions greater than 4 bits, the impact of compression on instability is minimal.

\subsection{Joint Effect of Dimension and Precision}
\label{sec:joint}
We study the effect of dimension and precision together, and show that overall, as the memory increases, the downstream instability decreases. We also propose a simple rule of thumb relating the memory and instability, and evaluate the relative impact of dimension and precision on the instability. Finally, we discuss two key questions based on our empirical observations, which motivate the rest of the work.

\vspace{-1.5mm}
\paragraph*{Tradeoffs}
We uniformly quantize the Wiki'17 and Wiki'18 embeddings of dimensions \{25, 50, 100, 200, 400, 800\} to precisions \{1, 2, 4, 8, 16, 32\} to generate many dimension-precision pairs spanning over a wide range of memory budgets. Across the memory budgets, embedding algorithms, and tasks, we see that as we increase the memory, the downstream instability decreases (Figure~\ref{fig:joint}). To propose a simple rule of thumb for the stability-memory tradeoff, we fit a single linear-log model to the dimension-precision pairs for all memory budgets less than $10^3$ bits/word (after which the instability plateaus) across five downstream tasks (i.e., the four sentiment analysis tasks and one NER task) and two embedding algorithms. We find the following average stability-memory relationship for the downstream instability $\mathcal{DI_T}$ for a task $T$ with respect to the memory, or bits/word, $M$ : $\mathcal{DI_T} \approx C_T - 1.3 * log_{2}(M)$, where $C_T$ is a task-specific constant. For instance, if we increase the memory \yell{$2\times$}, then the instability decreases on average by \yell{1.3\%} (absolute).
Across the tasks, embedding algorithms, and memory budgets we consider, this 1.3\% (absolute) difference corresponds to an approximately 5\% to 37\% \emph{relative} reduction in downstream instability, depending on the original instability value (3.6\% to 25.9\%).

To understand the relative impact on instability of increasing the dimension vs. the precision, we fit independent linear-log models to each parameter.
We find that precision has a larger impact on instability than dimension, with a \yell{2$\times$} increase in precision decreasing instability by \yell{1.4\%} (absolute) vs. a \yell{2$\times$} increase in dimension decreasing instability by \yell{1.2\%} (absolute). Please see Appendix~\ref{sec:linear-log-model} for more details on how we fit these trends. \emph{In Appendix~\ref{sec:robustness}, we further demonstrate the robustness of the stability-memory tradeoff (e.g., to more complex downstream models, other sources of downstream randomness)}.

This stability-memory tradeoff raises two key questions: (1) how can we theoretically explain this tradeoff between the embedding memory and the downstream stability, and (2) how can we jointly select the embedding's dimension-precision parameters to minimize the downstream instability? Practically, choosing these parameters is important, because downstream instability can vary over \yell{3\%} across the different combinations of dimension and precision for a given memory budget (Figure~\ref{fig:joint}). The goal of the remainder of the paper will be to shed light on these questions.

%% file: measuring.tex
\section{Analyzing Embedding Instability}
\label{sec:measuring}
To address both questions raised above, we present a new measure of embedding instability, the eigenspace instability measure, which we show is both theoretically and empirically related to the downstream instability of the embeddings.
The goal of this measure is to efficiently estimate, given two embeddings, how different the predictions of models trained with these embeddings will be. We first define the eigenspace instability measure and present its theoretical connection with downstream instability in Section~\ref{sec:sym_overlap}; we then propose using this measure to efficiently select parameters to minimize downstream instability in Section~\ref{sec:opt-proposal}.

\vspace{-2mm}
\subsection{Eigenspace Instability Measure}
\label{sec:sym_overlap}

We now define the eigenspace instability measure between two embeddings, and show that this measure is directly related to the expected disagreement between linear regression models trained using these embeddings.

\begin{definition}
Let $X=USV^T \in \RR^{n\times d}$ and $\tX=\tU\tS\tV^T\in \RR^{n\times k}$ be the singular value decompositions (SVDs) of two embedding matrices $X$ and $\tX$, and let $\Sigma \in \RR^{n \times n}$ be a positive semidefinite matrix.
Then the \textit{eigenspace instability measure} between $X$ and $\tX$, with respect to $\Sigma$, is defined as
\begin{eqnarray*}
    \mathcal{EI}_{\Sigma}(X,\tX) \!\defeq\! \frac{1}{\tr(\Sigma)}\tr\!\bigg(\!\!\!\left(\!UU^T \!\!+\! \tU\tU^T \!\!-\! 2 \tU\tU^T UU^T \!\right)\!\Sigma\!\bigg).
\end{eqnarray*}
\end{definition}

\vspace{-1.5mm}
Intuitively, this measure captures how different the subspaces spanned by the left singular vectors of $X$ and $\tX$ are to one another; the measure will be equal to zero when the left singular vectors of $X$ and $\tX$ span identical subspaces of $\RR^n$, and will be equal to one when these singular vectors span orthogonal subspaces of $\RR^n$ whose union covers the whole space.
We note that the left singular vectors are particularly important in the case of linear regression models, because the predictions of the learned model on the training examples depend only on the label vector and the left singular vectors of the data matrix.\footnote{The linear model trained on data matrix $X=USV^T \in \RR^{n\times d}$ with label vector $y \in \RR^n$ makes predictions $Xw = X(X^T X)^{-1} X^T y = UU^T y \in \RR^n$ on the $n$ training points.}

We now present our result showing that the expected mean squared difference between the linear regression models trained on $X$ vs.\ $\tX$  is equal to the the eigenspace instability measure, where $\Sigma$ corresponds to the covariance matrix of the regression label vector.
For the proof, see Appendix~\ref{app:theory}.

\begin{proposition}
\label{prop1}
Let $X\in \RR^{n\times d}$, $\tX\in \RR^{n\times k}$ be two full-rank embedding matrices, where $x_i$ and $\tx_i$ correspond to the $i^{th}$ rows of $X$ and $\tX$ respectively.
Let $y\in\RR^n$ be a random regression label vector with zero mean and covariance $\Sigma \in \RR^{n\times n}$.
Then the (normalized) expected mean squared difference between the linear models $f_y$ and $\tf_y$\footnote{$f_y(x) = w^T x$, for $w = (X^T X)^{-1}X^T y$, and $\tf_y(\tx) = \tw^T \tx$, for $\tw = (\tX^T \tX)^{-1}\tX^T y$.} trained on label vector $y$ using embeddings $X$ and $\tX$ satisfies
\begin{equation}
\frac{\expect{y}{\sum_{i=1}^n (f_y(x_i) - \tf_y(\tx_i))^2}}{\expect{y}{\|y\|^2}} =  \mathcal{EI}_{\Sigma}(X,\tX).
\end{equation}
\end{proposition}
\vspace{-2mm}
The above result exactly characterizes the expected downstream instability of linear regression models trained on $X$ and $\tX$, in terms of the eigenspace instability measure, given the covariance matrix $\Sigma$ of the label vector; but how should we select $\Sigma$?
One desirable property for $\Sigma$ could be that it produce label vectors with higher variance in directions believed to be important, for example because they correspond to eigenvectors with large eigenvalues of an embedding's Gram matrix.
In Section~\ref{sec:experiments}, where we evaluate the instability of pairs of embeddings of various dimensions and precisions, we consider $\Sigma = (EE^T)^{\alpha} + (\tE\tE^T)^{\alpha}$; in those experiments, $E$ and $\tE$ are the highest-dimensional ($d=800$), full-precision embeddings for Wiki'17 and Wiki'18, respectively, and $\alpha$ is a scalar controlling the relative importance of the directions of high eigenvalue.
This choice of $\Sigma$ results in label vectors with large variance in the directions of high eigenvalues of these embedding matrices.
In Section~\ref{sec:corr} we show that when $\alpha$ is chosen appropriately, there is strong empirical correlation between the eigenspace instability measure
(with this $\Sigma$) and downstream instability.

\vspace{-2mm}
\subsection{Jointly Selecting Dimension and Precision}
\label{sec:opt-proposal}

We now demonstrate a practical utility of the eigenspace instability measure: we propose using the measure to efficiently select embedding dimension-precision parameters to minimize downstream instability without training the downstream models.
In particular, we propose an algorithm that takes two or more pairs of embeddings with different dimension-precision parameters as input, and outputs the pair with the lowest eigenspace instability measure between embeddings.
In Section~\ref{sec:selection}, we evaluate the performance of this proposed selection algorithm in two settings: first, a simple setting where the goal is to select the pair with the lowest downstream instability out of two randomly selected pairs, and second, a more challenging setting where the goal is to select the pair with the lowest downstream instability out of two or more pairs with the same memory budget.
In both settings, we demonstrate that the eigenspace instability measure outperforms the majority of embedding distance measures and is competitive with the other top-performing embedding distance measure, the k-NN measure.
\vspace{-2mm}

%% file: evaluation.tex
\section{Experiments}
\label{sec:experiments}
We now empirically validate the eigenspace instability measure's relation with downstream instability and demonstrate that the eigenspace instability measure is an effective selection criterion for dimension-precision parameters. In Section~\ref{sec:corr}, we show that the theoretically grounded eigenspace instability measure strongly correlates with downstream instability, attaining Spearman correlations greater than the weaker baselines (semantic displacement, PIP loss, and eigenspace overlap score) and between \yell{0.04} better and \yell{0.09} worse than the strongest baseline (the k-NN measure). In Section~\ref{sec:selection}, when selecting dimension-precision parameters without training the downstream models, we show that the eigenspace instability measure attains up to \yell{$3.33\times$} lower error rates than weaker baselines and from \yell{$0.95\times$ to $1.55\times$} the error rate of the k-NN measure.\footnote{Our code is available at \url{https://github.com/HazyResearch/anchor-stability}.}

\begin{table*}[t]
    \caption{Spearman correlation scores between embedding distance measures and downstream prediction disagreement across varying dimension-precision pairs for the embedding. Downstream models are trained for sentiment analysis (SST-2, Subj) and NER (CoNLL-2003) tasks. Strongest correlation values are bolded.}
    \medskip
    \small
    \label{tab:corr-scores}
    \begin{tabularx}{\textwidth}{lYYYYYYYYY}
    \toprule
    \textit{Downstream Task}  & \multicolumn{3}{c}{SST-2}  & \multicolumn{3}{c}{Subj} &  \multicolumn{3}{c}{CoNLL-2003} \\
    \midrule
    \textit{Embedding Algorithm}                          & CBOW  & GloVe & MC  & CBOW & GloVe & MC   & CBOW  & GloVe & MC  \\
    \cmidrule(lr){0-0}
    \cmidrule(lr){2-4}
    \cmidrule(lr){5-7}
    \cmidrule(lr){8-10}
    Eigenspace Instability   & 0.68	& 0.84	& 0.84	& 0.72	& \bf{0.77}	& \bf{0.78} & \bf{0.80} & 0.78 & 0.83 \\
    $1 - \text{k-NN}$        & \bf{0.74}	& \bf{0.86}	& \bf{0.89}	& \bf{0.74}	& 0.76	& 0.76 & 0.76 & \bf{0.86} & \bf{0.92} \\
    Semantic Displacement    & 0.70	& 0.34	& 0.28	& 0.45	& 0.43	& 0.46 & 0.53 & 0.16 & 0.32 \\
    PIP Loss                 & -0.40	& -0.06	& 0.39	& -0.14	& -0.14	& 0.56 & 0.01 & 0.11 & 0.44 \\
    $1-\text{Eigenspace Overlap}$ & 0.63	& 0.18	& 0.26	& 0.50	& 0.29	& 0.45 & 0.58 & 0.01 & 0.31 \\
    \bottomrule
    \end{tabularx}
    \end{table*}

\vspace{-1em}
\paragraph*{Experimental Setup}
\label{sec:exp-setup}
To evaluate how predictive the various embedding distance measures are of downstream instability, we take the embedding pairs and corresponding downstream models we trained in Section~\ref{sec:tradeoffs} and measure the embedding distance measures between these pairs of embeddings.
Specifically, we compute the k-NN measure, semantic displacement, PIP loss, eigenspace overlap score, and eigenspace instability measure between the embedding pairs (Section~\ref{sec:baselines}). Recall that the k-NN measure and the eigenspace instability measure each have an important hyperparameter: the $k$ in the k-NN measure, which determines how many neighbors we compare, and the $\alpha$ in the eigenspace instability measure, which controls how important the eigenvectors of high eigenvalue are. For both hyperparameters, we choose the values with the highest average correlation across four sentiment analysis tasks (SST-2, MR, Subj, and MPQA) and one NER task (CoNLL-2003) and two embedding algorithms (CBOW and MC)\footnote{These values also worked well for GloVe (added later).} when using validation datasets for the downstream tasks (\yell{$k$ = 5} and \yell{$\alpha$ = 3}). See Appendix~\ref{sec:measure-hyperparams} for more details. The eigenspace instability measure also requires additional embeddings $E$ and $\tilde{E}$: we use 800-dimensional, full-precision Wiki'17 and Wiki'18 embeddings as these are the highest dimensional, full-precision embeddings in our study.

\label{sec:exp}
\vspace{-1mm}
\subsection{Predictive Performance of the Eigenspace Instability Measure}
\label{sec:corr}
We evaluate how predictive the eigenspace instability measure is of downstream instability, showing that the theoretically grounded eigenspace instability measure correlates strongly with downstream instability and is competitive with other embedding distance measures. To do this, we measure the Spearman correlations between the downstream prediction disagreement and the embedding distance measure for each of the five tasks and three embedding algorithms.
The Spearman correlation quantifies how similar the ranking of the pairs of embeddings based on the embedding distance measure is to the ranking of the pairs of embeddings based on their downstream prediction disagreement, with a maximum value of 1.0.
In Table~\ref{tab:corr-scores}, we see that the eigenspace instability measure and the k-NN measure are the top-performing embedding distance measures by Spearman correlation, with the eigenspace instability measure attaining Spearman correlations between \yell{0.04} better and \yell{0.09} worse than the k-NN measure on all tasks. Moreover, the strong correlation of at least \yell{0.68} for the eigenspace instability measure across embedding algorithms and downstream tasks validates our theoretical claim that this measure relates to downstream disagreement. In Appendix~\ref{sec:app-corr}, we include additional plots showing the downstream prediction disagreement versus the embedding distance measures.

\begin{table*}[t]
    \caption{Selection error when using embedding distance measures to predict the most stable embedding dimension-precision parameters on sentiment analysis (SST-2, Subj) and NER (CoNLL-2003) downstream tasks. Lowest errors are bolded.}
    \medskip
    \small
    \label{tab:sel-scores}
    \begin{tabularx}{\textwidth}{lYYYYYYYYY}
    \toprule
    \textit{Downstream Task}  & \multicolumn{3}{c}{SST-2}  & \multicolumn{3}{c}{Subj} &  \multicolumn{3}{c}{CoNLL-2003} \\
    \midrule
    \textit{Embedding Algorithm}                          & CBOW  & GloVe & MC  & CBOW & GloVe & MC   & CBOW  & GloVe & MC  \\
    \cmidrule(lr){0-0}
    \cmidrule(lr){2-4}
    \cmidrule(lr){5-7}
    \cmidrule(lr){8-10}
    Eigenspace Instability   & 0.23	& 0.15	& 0.17	& 0.24	& \bf{0.21}	 & \bf{0.20}	& \bf{0.20}	& 0.20	& 0.17 \\
    $1 - \text{k-NN}$                     & \bf{0.21}	& \bf{0.14}	& \bf{0.13}	& \bf{0.23}	& \bf{0.21}	 & 0.21	& 0.21	& \bf{0.16}	& \bf{0.11} \\
    Semantic Displacement    & 0.24	& 0.40	& 0.42	& 0.34	& 0.36	 & 0.34	& 0.29	& 0.47	& 0.41 \\
    PIP Loss                 & 0.64	& 0.50	& 0.35	& 0.57	& 0.54	 & 0.28	& 0.50	& 0.44	& 0.32 \\
    $1-\text{Eigenspace Overlap}$     & 0.28	& 0.46	& 0.43	& 0.32	& 0.41	 & 0.34	& 0.29	& 0.52	& 0.41 \\
    \bottomrule
    \end{tabularx}
    \end{table*}

\begin{table*}[h]
    \caption{Average difference (absolute percentage) to the oracle downstream instability when using embedding distance measures as the selection criteria for dimension and precision parameters under fixed memory budgets. Top-performing values are bolded.
    }
    \medskip
    \small
    \label{tab:avg-case}
    \begin{tabularx}{\textwidth}{lYYYYYYYYY}
        \toprule
        \textit{Downstream Task}  & \multicolumn{3}{c}{SST-2}  & \multicolumn{3}{c}{Subj} &  \multicolumn{3}{c}{CoNLL-2003} \\
        \midrule
        \textit{Embedding Algorithm}                          & CBOW  & GloVe & MC  & CBOW & GloVe & MC   & CBOW  & GloVe & MC  \\
        \cmidrule(lr){0-0}
        \cmidrule(lr){2-4}
        \cmidrule(lr){5-7}
        \cmidrule(lr){8-10}
    Eigenspace Instability   & 0.65 &	    0.55    &	    1.42 &	    0.39 &	    \bf{0.41} &	0.63 &	0.28    &	\bf{0.45} &	0.43 \\
    $1 - \text{k-NN}$                     & 0.57 &	    \bf{0.43} &	    \bf{1.07} &	0.38 &	    0.44    & \bf{0.57} & 0.32 &	0.48 &	\bf{0.23} \\
    Semantic Displacement    & \bf{0.37} &	1.58 &	        3.73 &	    0.48 &	    0.64    &	0.94 &	0.27    &	0.89 &	1.17 \\
    PIP Loss                 & 3.63 &	    2.54 &	        3.32 &	    1.16 &	    1.71    &	0.74 &	0.83    &	0.83 &	0.99 \\
    $1-\text{Eigenspace Overlap}$     & 0.88 &	    1.58 &	        3.60 &	    \bf{0.34} &	0.64    &	0.93 & \bf{0.20} &	0.89 &	1.15 \\
    High Precision           & 0.85 &	    1.58 &	        3.94 &	    0.61 &	    0.64    &	1.01 &	0.60    &	0.89 &	1.28 \\
    Low Precision            & 3.63 &	    2.54 &	        1.23 &	    1.16 &	    1.71    &	1.39 &	0.83    &	0.83 &	0.74 \\
    \bottomrule
    \end{tabularx}
    \end{table*}

\vspace{-1mm}
\subsection{Embedding Distance Measures for Dimension-Precision Selection}
\label{sec:selection}
We demonstrate that the eigenspace instability measure is an effective selection criterion for dimension-precision parameters, outperforming the majority of existing embedding distance measures and competitive with the k-NN measure, for which there are no theoretical guarantees.
Specifically, we evaluate the embedding distance measures as selection criteria in two settings of increasing difficulty: in the first setting the goal is, given two pairs of embeddings (each corresponding to an arbitrary dimension-precision combination), to select the pair with the lowest downstream instability.
In the second, more challenging setting, the goal is to select, among all dimension-precision combinations corresponding to the same total memory, the one with the lowest downstream instability.
This setting is challenging, as for many memory budgets, there are more than two choices of embedding pairs, and some choices may have very similar expected downstream instability.
We now discuss each of these settings, and the corresponding results, in more detail.

For the first, simpler setting, we first form all groupings of two embedding pairs with different dimension-precision combinations. For instance, a grouping may have one embedding pair with dimension 800, precision 32, and another embedding pair with dimension 200, precision 2, where a pair consists of a Wiki'17 and a Wiki'18 embedding from the same algorithm. For each embedding distance measure, we report the fraction of groupings where the embedding distance measure correctly chooses the embedding pair with lower downstream instability on a given task.
We repeat over three seeds, comparing embedding pairs of the same seed, and report the average.
In Table~\ref{tab:sel-scores}, we show that the eigenspace instability measure and k-NN measure are the most accurate embedding distance measures, with up to \yell{3.33$\times$} and \yell{3.73$\times$} lower selection error rates than the other embedding distance measures, respectively.
Moreover, across downstream tasks, the eigenspace instability measure attains \yell{$0.95\times$ to $1.55\times$} the error rate of the k-NN measure.

For the second, more challenging setting, we enumerate all embedding pairs with different dimension-precision combinations \emph{which correspond to the same total memory}.
For each embedding measure, we report the average absolute percentage difference between the downstream instability of the pair selected by the measure to the most stable ``oracle" pair, across different memory budgets.
We also introduce two naive baselines that do not require an embedding distance measure: \emph{high precision}, which selects the pair with the highest precision possible at each memory budget, and \emph{low precision}, which selects the pair with the lowest precision possible at each memory budget.
As before, we repeat over three seeds, comparing embedding pairs of the same seed, and report the average.
We see that the eigenspace instability measure and k-NN measure again outperform the other baselines on the majority of downstream tasks, with the eigenspace instability measure attaining a distance up to \yell{2.98\%} (absolute) closer to the oracle than the other baselines, and average distance to the oracle \yell{0.03\%} (absolute) better to \yell{0.35\%} (absolute) worse than the k-NN measure across downstream tasks (Table~\ref{tab:avg-case}).
For both settings, we include additional results measuring the worst-case performance of the embedding distance measure in Appendix~\ref{sec:wc-robustness}, where we find that the eigenspace instability measure and k-NN measure continue to be the top-performing measures.

%% file: extensions.tex
\section{Extensions}
\label{sec:extensions}
We demonstrate that the stability-memory tradeoffs we observe with pre-trained word embeddings can extend to knowledge graph embeddings and contextual word embeddings: as the memory of the embedding increases, the instability decreases.
We first show how these trends hold on knowledge graph embeddings in Section~\ref{sec:kg} and then on contextual word embeddings in Section~\ref{sec:bert}.
\vspace{-2mm}
\subsection{Knowledge Graph Embeddings}
\label{sec:kg}
Knowledge graph embeddings (KGEs) are a popular type of embedding that is used for multi-relational data, such as social networks, knowledge bases, and recommender systems. Here, we show that as the dimension and precision of the KGE increases, the stability on two standard KGE tasks improves, aligning with the trends we observed on pre-trained word embedding algorithms. Unlike word embedding algorithms, the input to KGE algorithms is a directed graph, where the relations are the edges in the graph and the entities are the nodes in the graph. The graph can be represented as a set of triplets $(h,r,t)$, where the entity head $h$ is related by the relation $r$ to the entity tail $t$. The output is two sets of embeddings: (1) entity embeddings $(\mathbf{e}_h)$ and (2) relation embeddings $(\mathbf{r}_r)$. We study the stability of these embeddings on two standard benchmark tasks: link prediction and triplet classification. We summarize the datasets and protocols, and then discuss the results.

\vspace{-1mm}

\paragraph*{Datasets} We use two datasets to train KGE embeddings: FB15K-95 and FB15K. FB15K was introduced in \citet{bordes2013translating} and is composed of a subset of triplets from the Freebase knowledge base. We construct FB15K-95 by randomly sampling 95\% of the the triplets from the training dataset of FB15K. The validation and test datasets remain the same for both datasets. We use these datasets to study the stability of KGEs under small changes in training data.
\vspace{-2mm}
\paragraph*{Training Protocol}
We consider a standard KGE algorithm---TransE~\cite{bordes2013translating}. The TransE objective function minimizes the distances $d(\mathbf{e}_h+\mathbf{r}_r, \mathbf{e}_t)$ for observed triplets and maximizes the distances for negatively sampled triplets, where either $h$ or $t$ has been corrupted. We use the $L_{1}$ distance for the distance function $d$, and learn the embeddings iteratively via stochastic gradient descent.

To measure the impact of the dimension and precision on the stability of TransE embeddings, we train TransE embeddings of dimensions \{10, 20, 50, 100, 200, 400\} and then uniformly quantize the entity and relation embeddings for each TransE embedding to bits \{1, 2, 4, 8, 16, 32\} per entry in embedding.\footnote{The same dimension is used for both the entity and the relation embeddings.} We perform a hyperparameter sweep on the learning using dimension 50, and select the best learning rate on the validation set for link prediction. We use this learning rate for all dimensions to minimize the impact of learning rate on our analysis. We take other training hyperparameters from the TransE paper~\cite{bordes2013translating} for the FB15K dataset, and use three seeds to train each dimension using the OpenKE repository~\cite{han2018openke}.
\vspace{-2mm}
\paragraph*{Evaluation Protocol}
For each dimension-precision, we evaluate all pairs of embeddings trained on FB15K-95 and FB15K on the link prediction and triplet classification tasks.
For each test triplet, the link prediction task evaluates the mean predicted rank of an observed triplet among all corrupted triplets.
We measure instability on this task with unstable-rank@10: the fraction of changes in rank greater than 10 between two embeddings across all test triplets.

The triplet classification task was introduced in \citet{Socher2013ReasoningWN} and is a binary classification task to determine whether or not a triplet occurs in the knowledge graph. For each relation, a threshold $T_R$ is determined based on the validation set, such that if $d(\mathbf{e}_h+\mathbf{r}_r, \mathbf{e}_t) \leq T_R$ then the triplet is predicted as positive. For each dimension-precision pair, we set the thresholds on FB15K-95 embedding and use the same thresholds for the FB15K embedding. We include results with threshold set independently for each embedding in Appendix~\ref{sec:app-kg-exp}. As for classification with downstream NLP tasks, we define stability on the triplet classification task as the percentage prediction disagreement.
\vspace{-1.5mm}

\begin{figure}[t]
    \centering
    \includegraphics[width=\columnwidth]{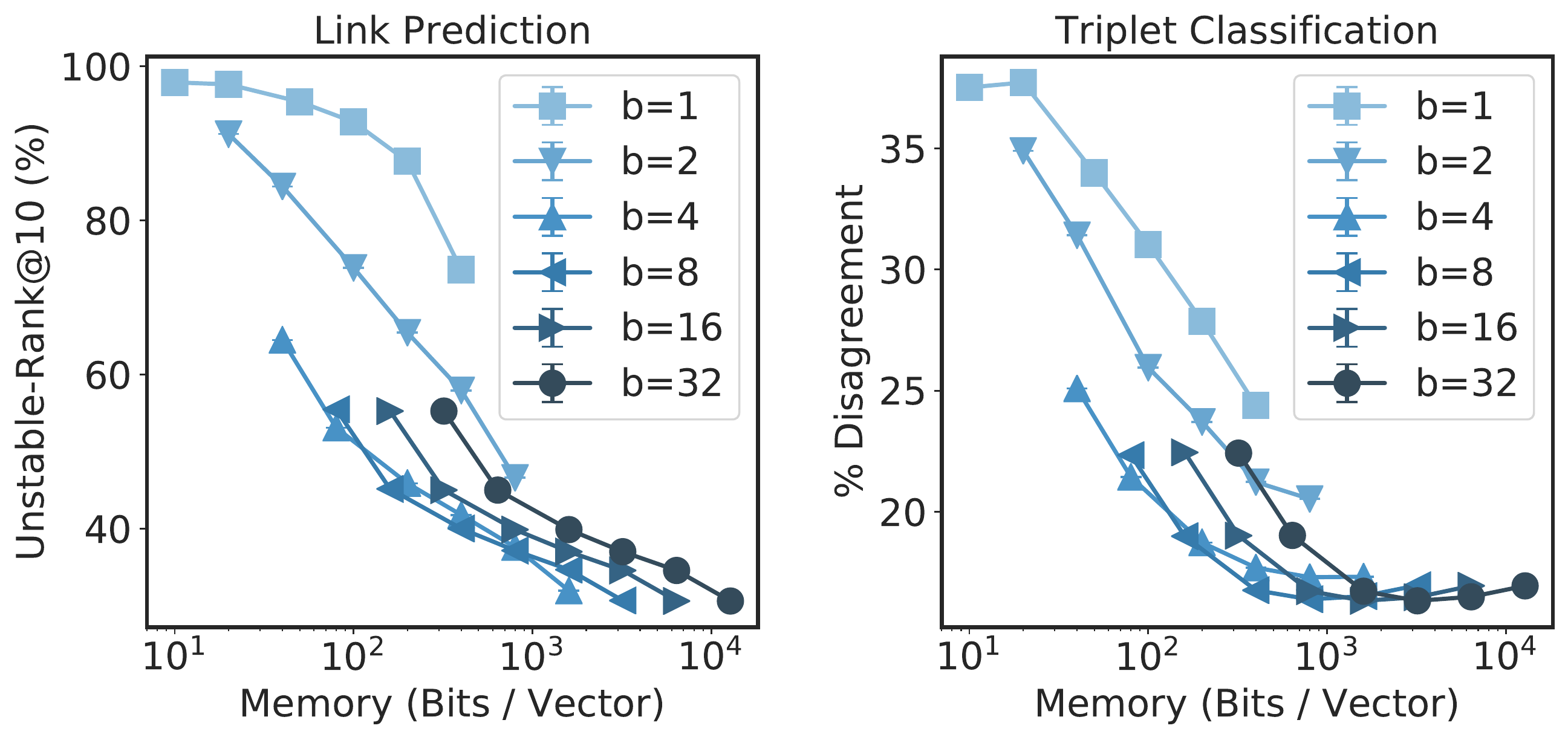}
    \caption{Stability of link prediction (left) and triplet classification (right) when evaluating embeddings trained on 95\% of FB15K training triplets and all of FB15K.}
    \vspace{-1em}
    \label{fig:transe}
    \end{figure}

\paragraph*{Results}
We find that the stability-memory tradeoffs continue to hold for TransE embeddings on the link prediction and triplet classification tasks: overall as the memory increases, the instability decreases, and specifically, as the dimension and precision increases, the instability decreases.
In Figure~\ref{fig:transe} (left), we show for link prediction that as the memory per vector increases, the unstable-rank@10 measure decreases.
Each line represents a different precision, where each point on the line represents a different dimension. Thus, we can also see that as the dimension increases, the unstable-rank@10 decreases, and as precision increases, this measure also decreases.
When fitting a linear-log model to the dimension-precision combinations for all memory budgets, we find that increasing the memory \yell{2$\times$} decreases the instability by \yell{7\% to 19\%} (relative).
In Figure~\ref{fig:transe} (right), we similarly show for triplet classification that as the memory per vector increases, the prediction disagreement between the embeddings trained on the two datasets decreases.
Finally, as we saw with word embeddings, we observe that the effect of the dimension or precision on stability is more significant at low memory regimes.

\vspace{-2mm}
\subsection{Contextual Word Embeddings}
\label{sec:bert}
Unlike pre-trained word embeddings, contextual word embeddings~\cite{peters2018deep,vaswani2017attention} extract word representations dynamically with awareness of the input context.
We find that the stability-memory trade-off observed on pre-trained embeddings can still hold for contextual word embeddings, though with noisier trends: higher dimensionality and higher precision can demonstrate better downstream stability.
We pre-train shallow, 3-layer versions of BERT~\cite{Devlin2018BERTPO} on sub-sampled Wiki'17 and Wiki'18 dumps ($\sim$200 million tokens) as feature extractors with different transformer layer output dimensionalities, ranging from a quarter as large to 4$\times$ as large as the hidden size in \texorpdfstring{BERT\textsubscript{BASE}}{BERT BASE} (i.e., 768).\footnote{The recent 12-layer \texorpdfstring{BERT\textsubscript{BASE}}{BERT BASE} model is pre-trained with 3 billion tokens from BooksCorpus~\cite{zhu2015aligning} and Wikipedia, and requires 16 TPU chips to train for 4 days.}
To evaluate the effect of precision, we use uniform quantization to compress the output of the last transformer layer in the BERT models.
Finally, we measure the prediction disagreement between linear classifiers trained on top of the Wiki'17 and Wiki'18 BERT models, with the BERT model parameters fixed.

Across four sentiment analysis tasks, we can observe reduced instability with higher dimensional BERT embeddings (Figure~\ref{fig:bert_stab_dim} in Appendix~\ref{app:bert}); however, the reduction in instability from increasing the dimension is noisier than with pre-trained word embeddings.
We hypothesize this is due to the instability of the training of the BERT embedding itself, which is a much more complex model than pre-trained word embeddings.
We also observe that increasing the precision can decrease the downstream instability, such that using 1 or 2 bits for precision often demonstrates observable degradation in stability, but precisions higher than 4-bit have negligible influence on stability (Figure~\ref{fig:bert_stab_prec} in Appendix~\ref{app:bert}).
For more details on the training and evaluation, see Appendix~\ref{app:bert}.

%% file: related.tex
\vspace{-1mm}
\section{Related work}
\label{sec:related}

There have been many recent works studying word embedding instability~\cite{hellrich2016bad,antoniak2018evaluating,wendlandt18surprising,Pierrejean2018PredictingWE,chugh,Hellrich2019TheIO}; these works have focused on the \emph{intrinsic} instability of word embeddings, meaning the stability measured between the embedding matrices without training a downstream model.
In the work of \citet{wendlandt18surprising} they do consider a downstream task (part-of-speech tagging), but focus on how the intrinsic instability impacts the \emph{error} of words on this task.
In contrast, we focus on the downstream instability (\ie, prediction disagreement), evaluating how different parameters of embeddings impact downstream instability with large-scale Wikipedia embeddings over multiple downstream NLP tasks. Furthermore, we provide theoretical analysis which is specific to the downstream instability setting to help explain our empirical observations.

More broadly, researchers have also studied the general problem of ML model instability in the context of online training and incremental learning. \citet{Fard2016LaunchAI} study the problem of reducing the prediction churn between consecutively trained classifiers by introducing a Monte Carlo stabilization operator as a form of regularization. \citet{Cotter2016SatisfyingRG} further define stability as a design goal for classifiers in real-world applications, along with goals such as precision, recall, and fairness, and propose an algorithm to optimize for these multiple design goals. Other researchers have also studied the problem of catastrophic forgetting when models are incrementally trained~\cite{Yang2019AdaptiveDM}, which shares a similar goal of wanting to learn new information, while minimizing changes with respect to previous models. As these works focus on changes to the downstream model training to reduce instability, we believe these works are complementary to our work, which focuses on better understanding the instability introduced by word embeddings.

Lastly, although the bias-variance tradeoff is a commonly used tool in ML to analyze model stability, there is an important difference in our setting. While the variance of a model quantifies the expected deviation of the model from its mean (typically over randomness in the training sample), in our work we analyze the disagreement between two separate models trained with different fixed data matrices on the same random label vector.

%% file: conclusion.tex
\vspace{-1mm}
\section{Conclusion}

We performed the first in-depth study of the downstream instability of word embeddings. In our study, we exposed a novel stability-memory tradeoff, showing that increasing the embedding dimension or precision decreases downstream instability. To better understand these empirical results, we introduced a new measure for embedding instability---the eigenspace instability measure---which we theoretically relate to downstream prediction disagreement. We showed that this theoretically grounded embedding measure correlates strongly with downstream instability, and can be used to select dimension-precision parameters, performing better than or competitively with other embedding measures on minimizing downstream instability without training the downstream tasks. Finally, we demonstrated that the stability-memory tradeoff extends to other types of embeddings, including contextual word embeddings and knowledge graph embeddings. We hope our study motivates future work on ML model instability in more complex pipelines.

%% file: artifact.tex
\section{Artifact Appendix}

\subsection{Abstract}

This artifact reproduces the memory-stability tradeoff and the embedding distance measure results for the sentiment analysis experiments on the word2vec CBOW and matrix completion embedding algorithms. It contains the pre-trained CBOW and MC embeddings of six different dimensions, trained on the Wiki'17 and Wiki'18 datasets, and the scripts and data for training the sentiment analysis tasks on these embeddings. It can validate the results in Figures 1 and 2, and Tables 1, 2, and 3 for the sentiment analysis tasks for the MC and CBOW embedding algorithms. We describe the specific steps to reproduce the SST-2 sentiment analysis results (which received the ACM badges), however, the steps can be easily modified to validate the MR, Subj, and MPQA sentiment tasks.

Our experimental pipeline consists of 3 main steps: (1) train and compress embeddings, (2) train downstream models and compute metrics, and (3) run analyses. Because step (1) is very computationally expensive (takes approximately 600 CPU hours to train all CBOW and MC embeddings using 56 threads), we provide the pre-trained embeddings (they must still be compressed). This artifact supports reproducing steps (1), (2), and (3), starting from the compression of the embeddings. The full artifact requires 1.1 TB of disk space for storing all embeddings and model output, and requires at least 1 GPU (tested on NVIDIA K80s) for training downstream models. We also provide a lightweight option to start from (3), which does not require training downstream models or space to store embeddings and can be run on a local machine. To do this, we provide CSVs of the pre-computed embedding distance measures and downstream instabilities.

\subsection{Artifact check-list (meta-information)}

{\small

\begin{itemize}
  \item {\bf Model: } Linear bag-of-words model for sentiment analysis (included).
  \item {\bf Data set:}  SST-2\footnote{We describe how to modify the scripts for the other provided datasets of MR, Subj, MPQA in Section~\ref{sec:custom}.}(included).
  \item {\bf Run-time environment: } Debian GNU/Linux, or Ubuntu 16.04 with CUDA ($\geq$ 9.0).
  \item {\bf Hardware: } Compute node (Amazon EC2 p2.16xlarge or equivalent) with  at least 1 NVIDIA K80 for model training.
  \item {\bf Metrics: } Embedding distance measures and downstream instability (defined in Section 2).
  \item {\bf Output: } Reproduces SST-2 results in Figures 1 and 2, and in Tables 1, 2, and 3.
  \item {\bf Experiments: } Included shell scripts, Jupyter notebook for plotting.
  \item {\bf How much disk space required (approximately)?: } ~900 GB for storing all embeddings, ~200 GB for storing SST-2 model and analysis results.
  \item {\bf How much time is needed to prepare workflow (approximately)?:} 30 minutes for installing dependencies.
  \item {\bf How much time is needed to complete experiments (approximately)?: } 17 CPU hours for embedding compression, 43 GPU hours for model training, 15 CPU hours for metric computation, and 1-3 minutes for analysis. Note embedding compression, model training, and metric computation are easily parallelizable.
  \item {\bf Publicly available?: } Yes.
  \item {\bf Code licenses (if publicly available)?: } MIT License.
\end{itemize}

\subsection{Description}

\subsubsection{How to access}

Our source code is publicly available on GitHub: \url{https://github.com/HazyResearch/anchor-stability}. Pre-trained embeddings are currently stored in a publicly accessible Google Cloud storage bucket (script to download from the bucket is provided in the GitHub repository in \texttt{run\_get\_embs.sh}).

We also have the source code and pre-trained emebddings permanently available at \url{https://doi.org/10.5281/zenodo.3687120}, which obtained the ACM badges.

\subsubsection{Hardware dependencies}

We recommend an Amazon EC2 p2.16xlarge or equivalent for the embedding compression, model training, and metric computation steps. For a Base AMI, we suggest the Deep Learning AMI (Ubuntu 16.04) Version 26.0 (ami-025ed45832b817a35).

\subsubsection{Software dependencies}

We tested our implementation on Ubuntu 16.04 with CUDA 9.0. We recommend using a conda environment or Python virtualenv, and we provide a \texttt{requirements.txt} file with the Python dependencies. We tested our implementation with Python 3.6 and PyTorch 1.0.

\subsection{Installation}

Please see the \url{https://github.com/HazyResearch/anchor-stability/blob/master/README.md} file for detailed installation instructions and scripts.

\subsection{Experiment workflow}
\label{sec:expwork}

We provide shell scripts to run to reproduce each of the steps.\footnote{If resource limited, to only reproduce the analysis results, we provide the CSVs of the embedding distance measures between pairs of embeddings and downstream instabilities between pairs of corresponding models.} Here we summarize the workflow; please see the \texttt{README.md} for more detailed instructions and specific commands to run.
\begin{enumerate}
    \item Obtain the pre-trained MC and CBOW embeddings trained on Wiki'17 and Wiki'18 and compress all embeddings to precisions $b \in \{1, 2, 4, 8, 16, 32\}$.
    \item Train the downstream models on top of all of the compressed embeddings for the SST-2 task. After the models are done training, compute the embedding distance measures and downstream instability between pairs of embeddings trained on Wiki'17 and Wiki'18, and their corresponding pairs of models.
    \item Run the analysis script to evaluate the Spearman correlations of the embedding distance measures with the downstream instabilities, and the selection criterion results for the tasks described in Section~\ref{sec:selection}. Finally, graph the memory-stability tradeoff results with the Jupyter notebooks provided.
\end{enumerate}

\subsection{Evaluation and expected result}

Step 3 in Section~\ref{sec:expwork} should reproduce the results for the CBOW and MC embeddings for the SST-2 sentiment analysis task in Figures 1 and 2, as well as the results in Tables 1, 2, and 3, using the compressed embeddings, trained models, and measured instabilities generated in Steps 1 and 2. Note there might be slight variance in the k-NN results (+/- 0.03 for Spearman correlation and selection error).

Using our provided CSVs file (see the \texttt{results} directory), Step 3 should also reproduce the remaining analysis results for the MR, Subj, and MPQA sentiment analysis tasks, as well as the CoNLL-2003 NER task found in Table 1, 2, and 3, as well as the linear-log trends described in Section~\ref{sec:tradeoffs}.

\subsection{Experiment customization}
\label{sec:custom}
To reproduce the complete pipeline of results on the MR, Subj, and MPQA tasks, modify the \texttt{run\_models.sh} and \texttt{run\_collect\_results.sh} script to use the MC and CBOW learning rates (\texttt{MC\_LR} and \texttt{CBOW\_LR}) that we found from our grid search (Appendix~\ref{sec:sa-setup}) for the new task and update the \texttt{DATASET} variable to the new task. Then pass the new task name to \texttt{run\_analysis.sh} (e.g., with \texttt{bash run\_analysis.sh mr}).

In terms of extending the results, the pre-trained embeddings we provide could be used to train more models to further measure the impact of the embedding instability on downstream instability. New embedding distance measures could also be added to \texttt{anchor/embedding.py} and easily evaluated against the measures presented in this paper in terms of their correlation with downstream instability.

\subsection{Methodology}

Submission, reviewing and badging methodology:

\begin{itemize}
  \item \url{http://cTuning.org/ae/submission-20200102.html}
  \item \url{http://cTuning.org/ae/reviewing-20200102.html}
  \item \url{https://www.acm.org/publications/policies/artifact-review-badging}
\end{itemize}

%% file: theory.tex
\section{Eigenspace Instability: Theory}
\label{app:theory}
We present the proof of Proposition~\ref{prop1}, which shows that the expected prediction disagreement between the linear regression models trained on embedding matrices $X$ and $\tX$ is equal to the eigenspace instability measure between $X$ and $\tX$.

\begin{customprop}{1}
Let $X\in \RR^{n\times d}$, $\tX\in \RR^{n\times k}$ be two full-rank embedding matrices, where $x_i$ and $\tx_i$ correspond to the $i^{th}$ rows of $X$ and $\tX$ respectively.
Let $y\in\RR^n$ be a random regression label vector with zero mean and covariance $\Sigma \in \RR^{n\times n}$.
Then the (normalized) expected disagreement between the linear models $f_y$ and $\tf_y$\footnote{$f_y(x) = w^T x$, for $w = (X^T X)^{-1}X^T y$, and $\tf_y(\tx) = \tw^T \tx$, for $\tw = (\tX^T \tX)^{-1}\tX^T y$.} trained on label vector $y$ using embedding matrices $X$ and $\tX$ respectively satisfies
\begin{equation}
\frac{\expect{y}{\sum_{i=1}^n (f_y(x_i) - \tf_y(\tx_i))^2}}{\expect{y}{\|y\|^2}} =  \mathcal{EI}_{\Sigma}(X,\tX).
\end{equation}
\end{customprop}

\begin{proof}
Let $X=USV^T \in \RR^{n\times d}$ and $\tX=\tU\tS\tV^T \in \RR^{n\times d}$ be the SVDs of $X$ and $\tX$ respectively, and let $x_i$ and $\tx_i$ in $\RR^d$ be the $i^{th}$ rows of $X$ and $\tX$.
Recall that parameter vector $w \in \RR^d$ which minimizes $\|Xw-y\|_2^2$ is given by $w^* = (X^T X)^{-1} X^T y$ (where here we use the assumption that $X$ is full-rank to know that $X^TX$ is invertible).
Thus, the linear regression model $f_y(x)=x^T w^*$ trained on data matrix $X$ with label vector $y \in \RR^n$ makes predictions $Xw^* = X(X^T X)^{-1} X^T y = USV^T (VS^{-2}V^T)VSU^T y = UU^T y \in \RR^n$ on the $n$ training points.
So if we train linear model with data matrices $X$ and $\tX$, using the same label vector $y$, these model will make predictions $UU^Ty$ and $\tU\tU^Ty$ on the $n$ training points, respectively.
Thus, the expected disagreement between the predictions made using $X$ vs. $\tX$, over the randomness in $y$, can be expressed as follows:
\begin{eqnarray*}
	\expect{y}{\sum_{i=1}^n (f_y(x_i) - \tf_y(\tx_i))^2} &=& \expect{y}{\|UU^T y - \tU \tU^T y\|^2} \\
	&=& \expect{y}{\left(UU^T y - \tU \tU^T y\right)^T\left(UU^T y - \tU \tU^T y\right)} \\ 
	&=& \expect{y}{y^T UU^T UU^T y + y^T \tU\tU^T\tU\tU^T y - 2 y^T \tU\tU^T UU^T y} \\
	&=& \expect{y}{y^T \left(UU^T + \tU\tU^T - 2  \tU\tU^T UU^T \right)y} \\
	&=& \expect{y}{\tr\bigg(y^T \left(UU^T + \tU\tU^T - 2  \tU\tU^T UU^T \right)y\bigg)} \\
	&=& \tr\bigg(\left(UU^T + \tU\tU^T - 2  \tU\tU^T UU^T \right)\expect{y}{yy^T}\bigg) \\
	&=& \tr\bigg(\left(UU^T + \tU\tU^T - 2 \tU\tU^T UU^T \right)\Sigma\bigg), \quad \text{where } \Sigma= \expect{y}{yy^T}
\end{eqnarray*}

Furthermore, we can easily compute the expected norm of the label vector $y$.
\begin{eqnarray*}
\expect{y}{\|y\|^2} &=& \expect{y}{\tr(y^T y)} \\
&=& \expect{y}{\tr(yy^T)} \\
&=& \tr(\expect{y}{yy^T}) \\
&=& \tr(\Sigma).
\end{eqnarray*}
Thus, we have successfully shown that
\begin{eqnarray*}
\frac{\expect{y}{\sum_{i=1}^n (f_y(x_i) - \tf_y(\tx_i))^2}}{\expect{y}{\|y\|^2}} &=& \frac{\tr\bigg(\left(UU^T + \tU\tU^T - 2 \tU\tU^T UU^T \right)\Sigma\bigg)}{\tr(\Sigma)}\\
&\eqdef&  \mathcal{EI}_{\Sigma}(X,\tX),
\end{eqnarray*}
as desired.
\end{proof}

\subsection{Efficiently Computing the Eigenspace Instability Measure}
We now discuss an efficient way of computing the eigenspace instability measure, assuming $\Sigma = (EE^T)^{\alpha} + (\tE\tE^T)^{\alpha} = VR^{2\alpha}V^T + \tV \tR^{2\alpha}\tV^T$ as discussed in Section~\ref{sec:sym_overlap}.  Here, $E$ and $\tE$ correspond to fixed embedding matrices,\footnote{In our experiments, $E$ and $\tE$ are the highest-dimensional ($d=800$), full-precision embeddings for Wiki'17 and Wiki'18, respectively.} where $E = VRW^T$ and $\tE=\tV\tR\tW^T$ are the SVDs of $E$ and $\tE$ respectively.

Recall the definition of the eigenspace instability measure:
\begin{eqnarray*}
	\mathcal{EI}_{\Sigma}(X,\tX) \defeq \frac{1}{\tr(\Sigma)}\tr\bigg(\!\!\left(UU^T \!+\! \tU\tU^T \!-\! 2 \tU\tU^T UU^T \right)\Sigma\bigg).
\end{eqnarray*}
We now show that both traces in this expression can be computed efficiently.
\begin{eqnarray}
	\tr\bigg(\left(UU^T + \tU\tU^T - 2 \tU\tU^T UU^T \right)\Sigma\bigg)
	&=& \tr\bigg(\left(UU^T + \tU\tU^T - 2 \tU\tU^T UU^T \right)\left(VR^{2\alpha}V^T + \tV \tR^{2\alpha}\tV^T\right)\bigg) \nonumber \\
	&=& \tr\left(R^{\alpha} V^T UU^T V R^{\alpha}\right) + \tr\left(R^{\alpha} V^T \tU\tU^T V R^{\alpha}\right) - 2\tr\left(R^{\alpha} V^T \tU\tU^T UU^T V R^{\alpha}\right) + \nonumber \\
	&& \tr\left(\tR^{\alpha} \tV^T UU^T \tV \tR^{\alpha}\right) + \tr\left(\tR^{\alpha} \tV^T \tU\tU^T \tV \tR^{\alpha}\right) - 2\tr\left(\tR^{\alpha} \tV^T \tU\tU^T UU^T \tV \tR^{\alpha}\right) \nonumber \\
	&=& \|U^T V R^{\alpha}\|_F^2 + \|\tU^T V R^{\alpha}\|_F^2 - 2\tr\left(R^{\alpha} (V^T \tU)(\tU^T U)(U^T V) R^{\alpha}\right) + \nonumber \\
	&& \|U^T \tV \tR^{\alpha}\|_F^2 + \|\tU^T \tV \tR^{\alpha}\|_F^2 - 2\tr\left(\tR^{\alpha} (\tV^T \tU)(\tU^T U)(U^T \tV) \tR^{\alpha}\right). \label{tr1}\\
	\tr\left(\Sigma\right) &=& \tr\left(VR^{2\alpha}V^T + \tV \tR^{2\alpha}\tV^T\right) \nonumber  \\
	&=& \tr\left(V^TVR^{2\alpha}\right) + \tr\left(\tV^T\tV \tR^{2\alpha}\right)  \nonumber \\
	&=& \tr\left(R^{2\alpha}\right) + \tr\left(\tR^{2\alpha}\right). \label{tr2}
\end{eqnarray}
We now note that the traces in Equation~\eqref{tr2}, and all the matrix multiplications in Equation~\eqref{tr1} can be computed efficiently and with low-memory (no need to ever store an $n$ by $n$ Gram matrix, for example), assuming the embedding matrices are ``tall and thin'' (large vocabulary, relatively low-dimensional).
More specifically, the eigenspace instability measure can be computed in time $O(nd^2)$ and memory $O(d^2)$, where we take $X,\tX,E,\tE$ to all be in $\RR^{n \times d}$ (or in $\RR^{n \times d'}$ for $d' \leq d$).
Thus, even for large vocabulary $n$, the eigenspace instability measure can be computed relatively efficiently (assuming the dimension $d$ isn't too large).

%% file: extended_setup.tex
\section{Experimental Setup Details}
We discuss the experimental protocols used for each of our experiments. In Appendix~\ref{sec:we-train}, we discuss the training procedures for the word embeddings, and in Appendix~\ref{sec:compression-details}, we discuss how we compress and post-process the embeddings. In Appendix~\ref{sec:ds-setup}, we describe the models, datasets, and training procedures used for the downstream tasks in our study, and in Appendix~\ref{sec:linear-log-model}, we discuss how we analyze the instability trends we observe on these tasks. Finally, in Appendix~\ref{sec:kg-setup-app} and Appendix~\ref{sec:cwe-setup-app} we describe setup details for the extension experiments on knowledge graph and contextual word embeddings, respectively.

\subsection{Word Embedding Training}
\label{sec:we-train}
We use Google's C implementation of word2vec CBOW\footnote{\url{https://github.com/tmikolov/word2vec}}, the original GloVe implementation\footnote{\url{https://github.com/stanfordnlp/GloVe}}, and our own C++ implementation of MC to train word embeddings. For CBOW, we use the default learning rate. For GloVe, we use a learning rate of 0.01 (as the default of 0.05 resulted in NaNs on 800-dimensional Wiki embeddings). For MC, since we are using our own implementation, we use a learning rate which we found to achieve low loss on Wiki'17. We include the full details on the hyperparameters used for both embedding algorithms in Table~\ref{tab:emb-params}.

\begin{table}[h]
    \caption{Hyperparameters for embedding algorithms.}
    \medskip
    \small
    \label{tab:emb-params}
    \centering
    \begin{tabular}{llr}
    \toprule
    Algorithm & Hyperparameter & Value \\
    \midrule
    Shared         & Training epochs    & 50     \\
                   & Window size        & 15     \\
                   & Minimum count      & 5      \\
                   & Threads            & 56     \\
    \midrule
    CBOW           & Learning rate      & 0.05   \\
                   & Negative samples   & 5      \\
    \midrule
    GloVe          & Learning rate      & 0.01   \\
                   & $x_{max}$            & 100     \\
                   & $\alpha$           & 0.75   \\
    \midrule
    MC             & Learning rate      & 0.2    \\
                   & LR decay epochs    & 20     \\
                   & Batch size         & 128    \\
                   & Stopping tolerance & 0.0001 \\
    \bottomrule
    \end{tabular}
\end{table}

\subsection{Word Embedding Compression and Post-Processing}
\label{sec:compression-details}
We now discuss some important implementation details for uniform quantization related to stability. We use the techniques and implementation from \citet{smallfry}. To minimize confounding factors with stability, we use deterministic rounding for each word. The bounds of the interval for uniform quantization are determined by computing an optimal clipping threshold which is based on the distribution of the real numbers to be quantized. As we assume that embeddings $X$ and $\tilde{X}$ have similar distributions in terms of their vector values, we use the same clipping threshold across embeddings $X$ and $\tilde{X}$ to avoid unnecessary sources of instability, and we compute the clipping threshold using embedding $X$. Finally, we apply orthogonal Procrustes to align embedding $\tilde{X}$ to embedding $X$ before compressing the embeddings and training downstream models. Preliminary results indicated that this alignment decreased instability, particularly at high compression rates, and we use this technique throughout our experiments.

\subsection{Downstream Tasks}
\label{sec:ds-setup}
We discuss the models, datasets, and training procedure we use for the sentiment analysis and NER tasks.

\subsubsection{Sentiment Analysis}
\label{sec:sa-setup}
We use a simple, bag-of-words model for sentiment analysis. The goal of the task is to classify a sentence as positive or negative. For each sentence, the bag-of-words model averages the word embeddings of the words in the sentence and then passes the sentence embedding through a linear classifier. This simple model allows us to study the impact of the embedding on the downstream task in a controlled setting, where the downstream model itself is expected to be fairly stable.

We use four datasets for the sentiment analysis task: SST-2, MR, Subj, and MPQA. These are the four largest binary classification datasets used in ~\citet{Kim2014ConvolutionalNN}.\footnote{\url{https://github.com/harvardnlp/sent-conv-torch/tree/master/data}} We use their given train/validation/test splits for SST-2. For MR, Subj, and MPQA, which do not have these splits, we take 10\% of the data for the validation set, 10\% for the test set, and use the remaining 80\% for the training set.

We tune the learning rate for each dataset and embedding algorithm. We use the 400-dimensional Wiki'17 embeddings to tune the learning rate in the grid of \{1e-6, 1e-5, 0.0001, 0.001, 0.01, 0.1, 1\}. We choose the learning rate which achieves the highest validation accuracy on average across three seeds for each dataset and report the selected values in Table~\ref{tab:hyperparams-sent}. To avoid choosing unstable learning rates, we also throw out learning rate values where the validation errors increase by 15\% or greater between any consecutive epochs. We include the hyperparameters shared among all datasets in Table~\ref{tab:hyperparams-sent-all}.
\vspace{-3mm}
\begin{table}[h]
\centering
\caption{(a) shows selected learning rates on the sentiment analysis tasks for each embedding algorithm. (b) shows training hyperparameters for the sentiment analysis datasets that are shared across embedding algorithms.\vspace{3mm}}
\begin{subtable}{0.55\linewidth}
\centering
\caption{Tuned learning rates for the sentiment analysis datasets.}
\medskip
\small
\label{tab:hyperparams-sent}
\begin{tabular}{lrrrr}
\toprule
Algorithm & SST-2 & MR     & Subj  & MPQA  \\
\midrule
CBOW  & 0.0001 & 0.001 & 0.0001 & 0.001 \\
GloVe & 0.01   & 0.01  & 0.01   & 0.001 \\
MC    & 0.001  & 0.1   & 0.1    & 0.001 \\
\bottomrule
\end{tabular}
\end{subtable}
\begin{subtable}{0.4\linewidth}
    \centering
    \caption{Shared training hyperparameters.}
    \medskip
    \small
    \label{tab:hyperparams-sent-all}
    \begin{tabular}{lr}
    \toprule
        Hyperparameter       & Value \\
    \midrule
        Optimizer            & Adam   \\
        Batch size           & 32    \\
        Training epochs      & 100   \\
    \bottomrule
    \end{tabular}
\end{subtable}
\end{table}

\subsubsection{Named Entity Recognition}
We use the single-layer, BiLSTM model from \citet{Akbik2018ContextualSE} for named entity recognition.\footnote{\url{https://github.com/zalandoresearch/flair}} We turn off the conditional random field (CRF) for computational efficiency and include a smaller subset of results with the CRF turned on in Appendix~\ref{sec:complex-ds}.

We use the standard English CoNLL-2003 dataset with the default setup for dataset splits~\cite{tjongkimsang2003conll}. Following \citet{Gardner2017AllenNLP}, we ignore article divisions (denoted with ``-DOCSTART-") and do not consider them as sentences.\footnote{\url{https://github.com/allenai/allennlp}}

We tune the learning rate per embedding algorithm, and otherwise follow the training hyperparameter settings of \citet{Akbik2018ContextualSE}. Using the 400-dimensional Wiki'17 embeddings, we sweep the learning rate in the grid of \{0.001, 0.01, 0.1, 1, 10\}, and choose the one which achieves the highest validation micro F1-score on average across three seeds for each embedding algorithm. We train with vanilla SGD without momentum and use learning decay with early stopping if the learning rate becomes too small. We provide the selected learning rates in Table~\ref{tab:ner-params} and the hyperparameters shared across embeddings in Table~\ref{tab:ner-shared-params}.
\vspace{-4mm}
\begin{table}[h!]
    \centering
    \caption{(a) shows selected learning rates for the NER task per embedding algorithm. (b) shows training hyperparameters shared across embedding algorithms for the NER task.}
    \begin{subtable}{0.3\linewidth}
    \centering
    \caption{Tuned learning rates for NER.}
    \medskip
    \small
    \label{tab:ner-params}
    \begin{tabular}{ccc}
    \toprule
        CBOW & GloVe & MC \\
    \midrule
        0.1  & 1.0 &  1.0   \\
    \bottomrule
    \end{tabular}
\end{subtable}
\begin{subtable}{0.65\linewidth}
    \centering
\caption{Shared training hyperparameters.\vspace{-2mm}
}
\medskip
\small
\label{tab:ner-shared-params}
\begin{tabular}{lr}
\toprule
    Hyperparameter       & Value \\
\midrule
    Optimizer            & SGD   \\
    Batch size           & 32    \\
    Max. training epochs & 150   \\
    LSTM hidden size     & 256   \\
    LSTM num. layers     & 1     \\
    Patience             & 3     \\
    Anneal factor        & 0.5   \\
    Word dropout         & 0.05  \\
    Locked dropout       & 0.5   \\
\bottomrule
\end{tabular}
\end{subtable}
\end{table}

\subsection{Fitting Linear-Log Models to Trends}
\label{sec:linear-log-model}

We describe in detail how we fit linear-log model to the memory, dimension, and precision trends in Section~\ref{sec:joint}.
To propose the simple rule of thumb relating stability and memory, we consider 10 tasks to form a data matrix for the linear-log model: 5 downstream tasks (the four sentiment tasks in our study and the NER task) for two embedding algorithms (CBOW and MC embeddings).
Let $P$ denote the number of Wiki'17/Wiki'18 pairs of embedding matrices from our experiments which  correspond to a combination of dimension $d$, precision $b$, and random seed $s$  (we consider 3 random seeds) such that the number of bits per row is less than our cutoff of $10^3$ ($bd < 10^3$).\footnote{In our case $P=63$, because we have 3 random seeds, and 21 pairs of dimension $d\in\{25,50,100,200,400,800\}$ and precision $b\in\{1,2,4,8,16,32\}$ such that $db < 10^3$.}
For each task $t$ (out of $T=10$ total tasks), we construct a data matrix $X^{(t)} \in \RR^{P \times (T+1)}$, and a label vector $y^{(t)} \in \RR^P$, as follows:
Each row in $X^{(t)}$ corresponds to one of the above $P$ pairs of Wiki'17/Wiki'18 embedding matrices.
For each of these embedding matrix pairs, we compute the memory $m'$ in bits occupied per row of the embedding matrices, as well as the downstream prediction disagreement percentage $y' \in [0,100]$ between the models trained on those embeddings.
We then set the corresponding row in $X^{(t)}$ to be $[log_2{(m')}, e_t] \in \RR^{T+1}$, where $e_t \in \RR^{T}$ is a binary vector with a one at index $t$ and zeros everywhere else, and the corresponding entry of $y^{(t)}$ to the prediction disagreement $y'$; note that appending $e_t$ to $\log_2(m')$ allows us to learn a different bias term (\ie, $y$-intercept) per task.
We then vertically concatenate all the $X^{(t)}$ matrices and label vectors $y^{(t)}$, to form a single data matrix $X \in \RR^{TP \times (T+1)}$ and label vector $y \in \RR^{TP}$.
To fit our log-linear model, we use $X$ and $y$ to solve the least squares problem using the closed form solution, $\hat{\beta} = (X^T X)^{-1} X^T y$.
Given $\hat{\beta} \in \RR^{T+1}$, for each task $t$ we can extract the fitted log-linear trend:
$\mathcal{DI}_t \approx \hat{\beta}_t - \hat{\beta}_0 * log_{2}(m)$, where $\hat{\beta}_0 \approx 1.3$ is the first element of $\hat{\beta}$, and $\hat{\beta}_t$ is the $(t+1)^{th}$ element of $\hat{\beta}$.
This implies that doubling the memory of the embeddings on average leads to a 1.3\% reduction in downstream prediction disagreement.

To fit the individual dimension and precision log-linear trends, we follow a protocol very similar to the above.
For the dimension (respectively, precision) trend, the primary difference with the above protocol is that instead of having an independent $y$-intercept term per task, we have an independent $y$-intercept term for each combination of task and precision (resp., dimension).
Furthermore, in the rows of the data matrices, instead of $\log_2(\cdot)$ of the memory $m$, we consider $\log_2(\cdot)$ of the dimension $d$ (resp., precision $b$).

We also use the linear-log model for stability-memory to compute the minimum and maximum relative percentage decreases in downstream instability when increasing the memory of word embeddings.
In particular, our goal is to understand how much the 1.3\% decrease in prediction disagreement is in relative terms.
To do this, we consider the combination of downstream task and embedding algorithm which is most stable at high memory (task: Subj; embedding algorithm: CBOW), and the combination which is least stable at low memory (task: MR; embedding algorithm: MC).
At these extreme points, the instability is approximately 2.2\% and 25.9\%, respectively.
A  1.3\% absolute decrease in instability from 3.5\% to 2.2\% corresponds to a relative decrease of approximately 37\% $\left(\frac{1.3}{3.5} \approx 0.37 \right)$.
Similarly, a 1.3\% absolute decrease in instability from 25.9\% to 24.6\% corresponds to a relative decrease of approximately 5\% $\left(\frac{1.3}{25.9} \approx 0.05 \right)$.
Thus, we conclude that this 1.3\% absolute decrease in instability corresponds to a relative decrease in instability between 5\% and 37\%, across the tasks and embedding algorithms we consider.

We repeat the procedures above to fit a linear-log model to the stability-memory trend for knowledge graph embeddings in Section~\ref{sec:kg}.

\subsection{Knowledge Graph Embeddings}
\label{sec:kg-setup-app}
We use the OpenKE repository to generate knowledge graph embeddings~\cite{han2018openke}.\footnote{\url{https://github.com/thunlp/OpenKE/tree/OpenKE-PyTorch}} We follow the training hyperparameters described in \citet{bordes2013translating} for TransE embeddings for the FB15K dataset where available, and use default parameters from the OpenKE repository, otherwise. We modify the repository to follow the early stopping procedure and normalization of entity embeddings to follow the protocol of \citet{bordes2013translating}. We additionally sweep the learning rate in \{1e-5, 0.0001, 0.001, 0.01, 0.1\} using dimension 50 on the FB15K-95 dataset, and choose the learning rate which attains the lowest mean rank (i.e., highest quality) on the validation set for the link prediction task. We include the full hyperparameters in Table~\ref{tab:kge-params}. We also note that unlike with word embeddings, we do not align embeddings with orthogonal Procrustes before compressing the embeddings with uniform quantization. We found alignment to result in a quality drop on knowledge graph embeddings, likely due to the fact that there are two sets of embeddings jointly learned (relation and entity embeddings) which require more advanced alignment techniques.
\vspace{-3mm}
\begin{table}[h]
    \centering
    \caption{Hyperparameters for training TransE knowledge graph embeddings. Bolded values indicate we performed a grid search. Other values are from \citet{bordes2013translating} and \citet{han2018openke}.}
    \medskip
    \small
    \label{tab:kge-params}
    \begin{tabular}{lr}
        \toprule
        Hyperparameter                 & Value         \\
    \midrule
        Optimizer                      & SGD           \\
    Max. training epochs           & 1000          \\
    Num. batches                   & 100           \\
    Threads                        & 8             \\
    Early stopping patience        & 10            \\
    Head/tail replacement strategy & Uniform       \\
    Entity negative rate           & 1             \\
    Relation negative rate         & 0             \\
    Margin $\gamma$                & 1             \\
    Distance $d$                   & $L_{1}$ \\
    \textbf{Learning rate}                  & \textbf{0.001}   \\
    \bottomrule
    \end{tabular}
    \end{table}
 \vspace{-2mm}
\subsection{Contextual Word Embeddings}
\label{sec:cwe-setup-app}

To study the downstream instability of contextual word embeddings, we pre-train BERT~\citet{Devlin2018BERTPO} models and then use them as fixed feature extractors to train downstream task models. We use BERT without fine-tuning parameters for downstream tasks because our goal is to isolate and study the instability resulting from the difference in pre-training corpora; this is in analogy to our study in Section~\ref{sec:tradeoffs} on the instability of conventional fixed pre-trained embeddings.
\vspace{-2mm}
\paragraph{Pre-training} In the pre-training phase, we use Wikipedia dumps (the major component of the corpus used by~\citet{Devlin2018BERTPO}) to train the BERT models. We use Wiki'2017 and Wiki'2018 dumps respectively for pre-training to study the instability introduced by the change in corpora. We pre-train BERT models with 3 transformer layers on 10\% subsampled articles from the Wikipedia dumps, which consists of approximately 200 million tokens. We use these shallower BERT model on the subsampled pre-training corpus to allow for computationally feasible training of BERT models with different transformer output dimensionality. As our corpus size are different from the one used by the original BERT model~\cite{Devlin2018BERTPO}, we first grid search the pre-training learning rate with the subsampled Wiki'17 corpus using the same transformer output dimensionality as the BERT\textsubscript{BASE}. We then use the grid-searched optimal learning rate to pre-train the BERT model with different transformer dimensionality for both Wiki'17 and Wiki'18 corpus.\footnote{We follow the experiment design from pre-trained word embeddings to use the same learning rate for pre-training BERT models with transformer configurations.}
\vspace{-2mm}
\paragraph{Downstream Evaluation}
To evaluate the downstream instability of pre-trained BERT models, we take BERT model pairs with the same model configuration but trained on Wiki'17 and Wiki'18 respectively. We measure the percentage of disagreement in downstream task prediction of the BERT pairs as proxy for downstream instability. Specifically, we evaluate the instability on the sentiment analysis task using the SST, Subj, MR and MPQA datasets. In these tasks, we use linear  bag-of-words models on top of the last transformer layer output; this output acts as the contextual word vector representation. To train the sentiment analysis task models, we first grid-search the learning rate using BERT with 768-dimensional transformer output for each dataset and choose the value with the highest validation accuracy.\footnote{We use the dimensionality used for original BERT\textsubscript{BASE}~\cite{Devlin2018BERTPO}.} We then use the grid-searched learning rate to train the sentiment analysis models using different pre-trained BERT models. To ensure statistically meaningful results, we use three random seeds to pre-train BERT models and train the downstream sentiment analysis models. We otherwise use the same hyperparameters reported in Table~\ref{tab:hyperparams-sent-all}.

%% file: extended_empirical.tex
\section{Extended Empirical Results}
\label{sec:ext-emp}

We now present additional experimental results to further validate the claims in this paper and provide deeper analysis of our results. We organize this section as follows:
\begin{itemize}
    \item In Appendix~\ref{sec:ext-trends}, we present additional results showing that the stability-memory trends (and individual dimension and precision trends) hold on sentiment analysis tasks.
    \item In Appendix~\ref{sec:quality-tradeoffs}, we evaluate another important property--quality---exploring the tradeoffs of quality with memory and stability for the tasks in our study.
    \item In Appendix~\ref{sec:measure-hyperparams}, we discuss how we choose the additional hyperparameters required for both the k-NN measure and the eigenspace instability measure.
    \item In Appendix~\ref{sec:app-corr}, we use visualizations to further analyze the relationship between the downstream instability and the embedding distance measures.
    \item In Appendix~\ref{sec:wc-robustness}, we include additional results on sentiment analysis tasks for the evaluation of the embedding distance measures. We also evaluate the worst-case performance of the embedding distance measures as selection criteria, showing that the eigenspace instability measure and k-NN measure remain the top-performing measures overall.
    \item In Appendix~\ref{sec:app-kg-exp}, we experiment with a modified setup for the triplet classification task, showing that the trends continue to hold, but the instability plateaus faster under this modification.
    \item In Appendix~\ref{sec:app-cwe-exp}, we include the figures for the contextual word embedding results presented in Section~\ref{sec:bert}.
\end{itemize}

\subsection{Stability-Memory Tradeoff}
\label{sec:ext-trends}

\begin{figure*}[h]
    \centering
    \begin{subfigure}{\textwidth}
        \centering
        \includegraphics[width=0.8\textwidth]{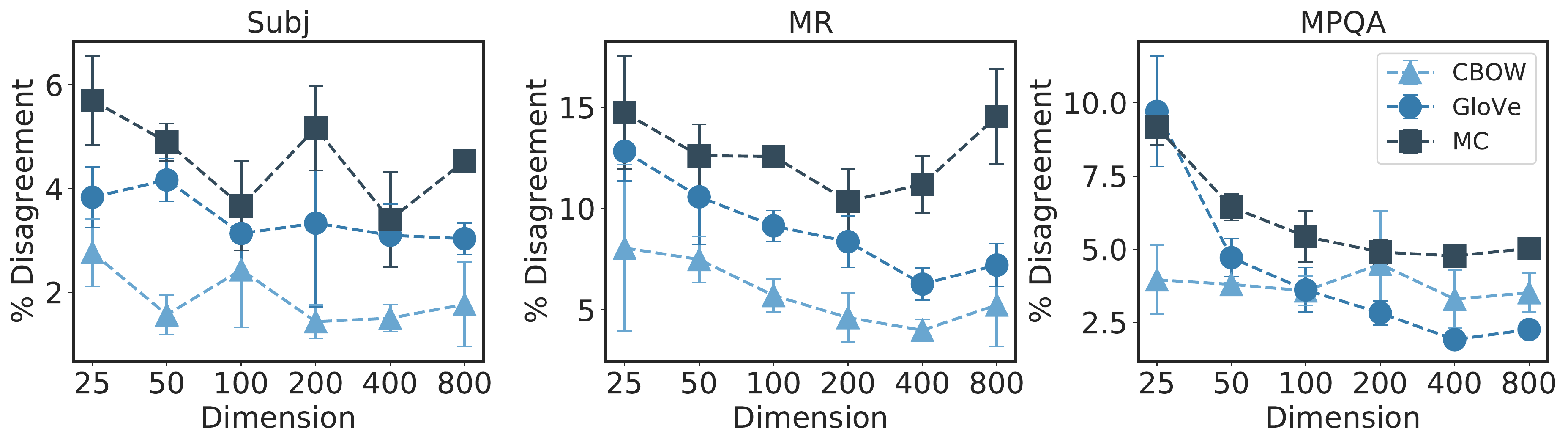}
    \caption{32-bit Precision}
    \end{subfigure}
    \begin{subfigure}{\textwidth}
        \centering
        \includegraphics[width=0.8\textwidth]{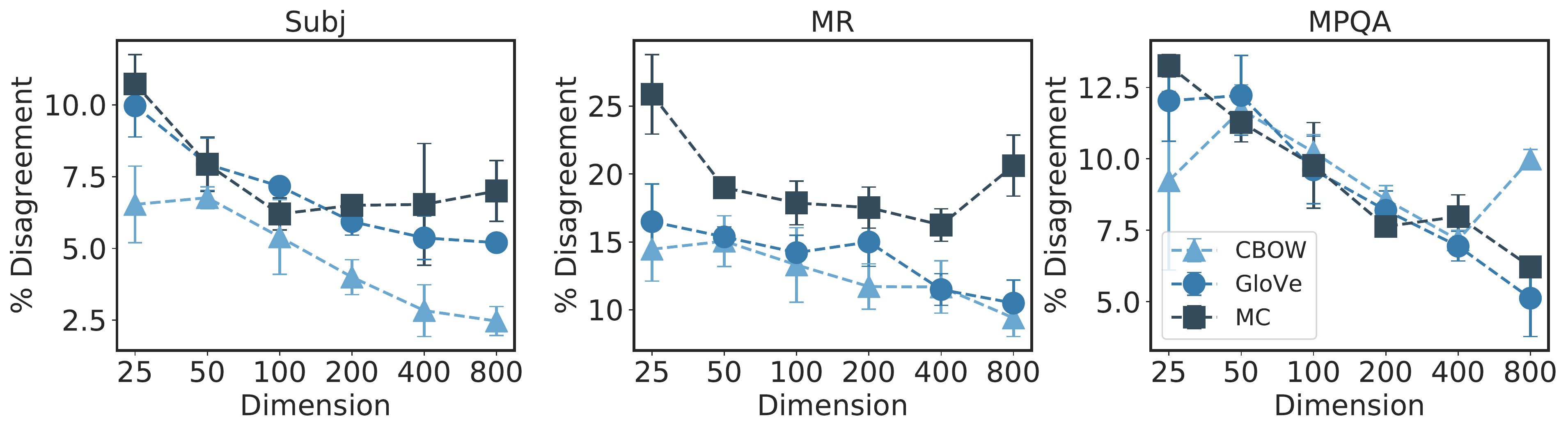}
    \caption{1-bit Precision}
    \end{subfigure}
    \caption{The effect of embedding dimension on the downstream instability of sentiment analysis tasks for CBOW, GloVe, and MC. We show the results at two different precisions: (top) 32-bit precision (uncompressed), and (bottom) 1-bit precision (32$\protect\times$ compressed).\vspace{-1mm}}
    \label{fig:ext-dim}
\end{figure*}

\begin{figure*}[h]
    \centering
        \includegraphics[width=0.8\textwidth]{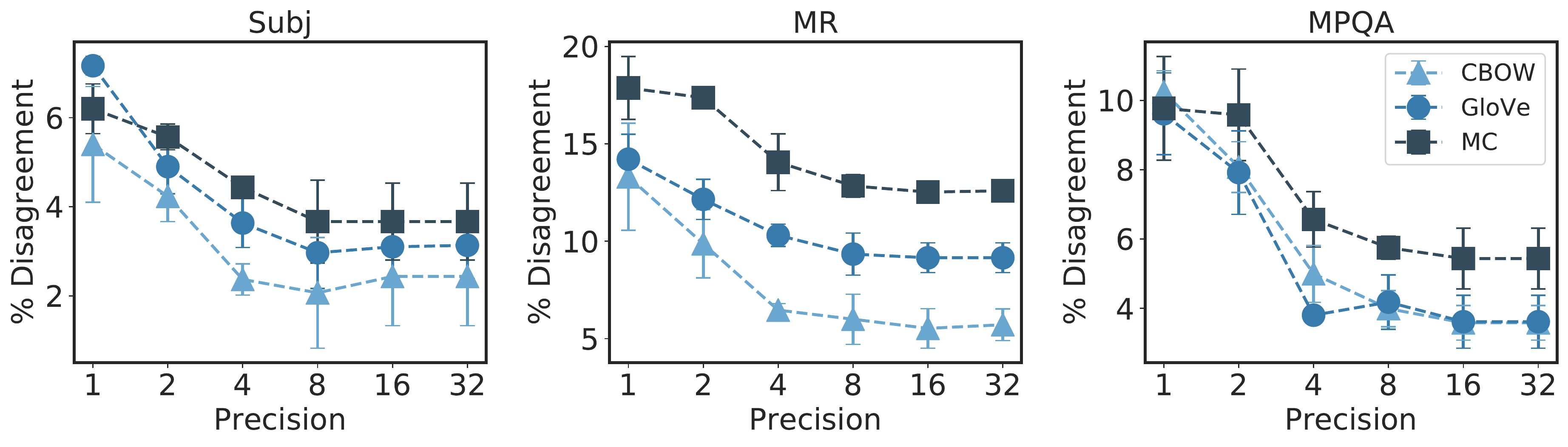}
    \caption{The effect of embedding precision on the downstream instability of sentiment analysis tasks for CBOW, GloVe, and MC embedding algorithms with 100-dimensional embeddings.\vspace{-3mm}}
    \label{fig:ext-prec}
\end{figure*}

\begin{figure*}[!h]
    \centering
        \includegraphics[width=\textwidth]{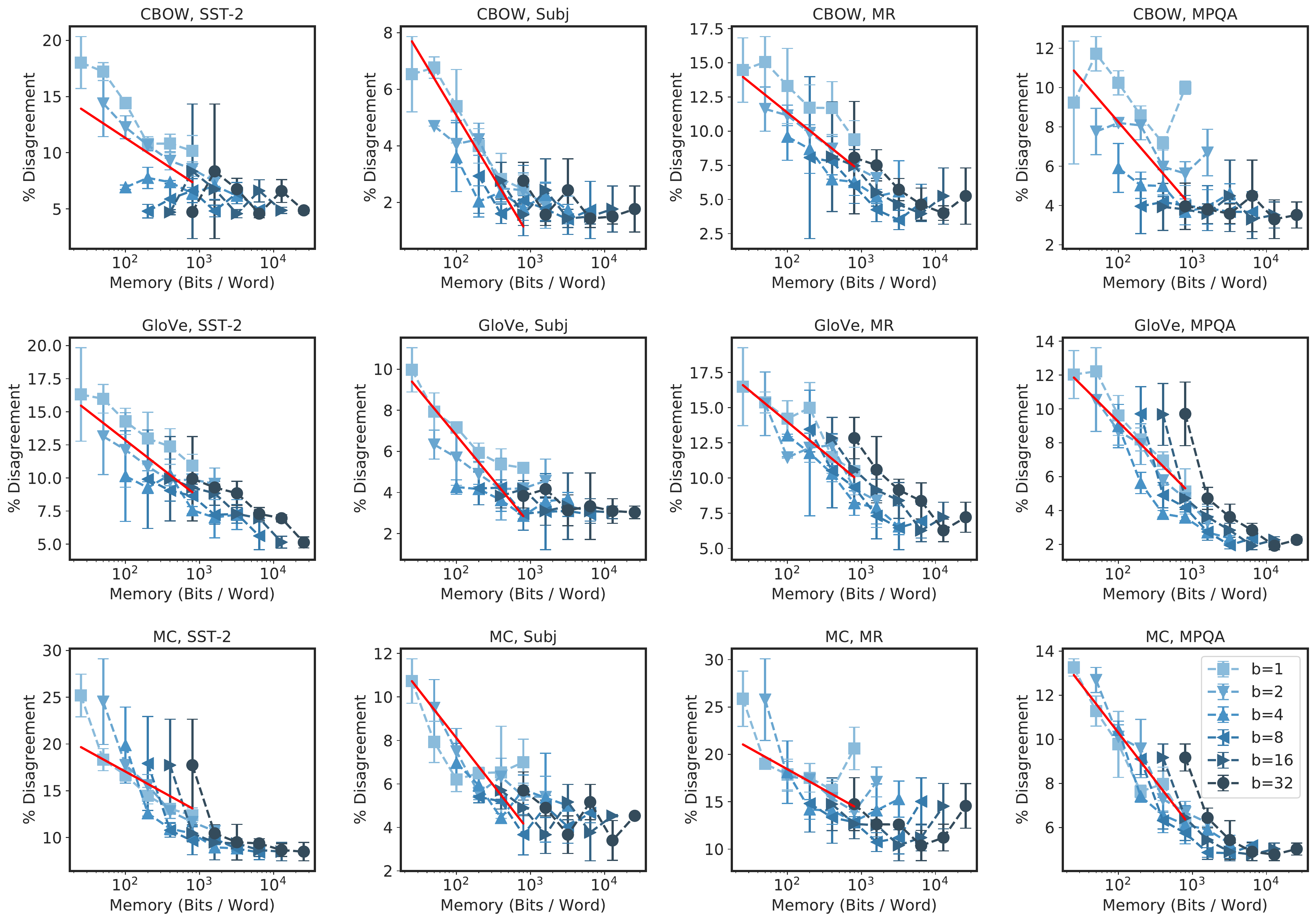}
    \caption{The effect of embedding dimension and precision on the downstream instability of sentiment analysis tasks for CBOW, GloVe, and MC embedding algorithms.\vspace{-2mm}}
    \vspace{-1em}
    \label{fig:ext-joint}
\end{figure*}

We validate that the stability-memory tradeoff holds on three more sentiment tasks (Subj, MR, and MPQA) for dimension and precision, first in isolation and then together. As always, we train embeddings and downstream models over three seeds, and the error bars indicate the standard deviation over these seeds.
In Figure~\ref{fig:ext-dim}, we can see more evidence that as the dimension increases, the downstream instability often decreases, with the trends more consistent for lower precision embeddings.
In Figure~\ref{fig:ext-prec}, we further validate that as the precision increases, the downstream instability decreases.
Finally, in Figure~\ref{fig:ext-joint}, we show on all four sentiment tasks (SST-2, Subj, MR, and MPQA) that when jointly varying dimension and precision, the instability decreases as the memory increases.

\vspace{-2mm}
\subsection{Quality Tradeoffs}
\label{sec:quality-tradeoffs}
\begin{figure*}[!h]
    \centering
    \begin{subfigure}{\textwidth}
    \centering
        \includegraphics[width=\textwidth]{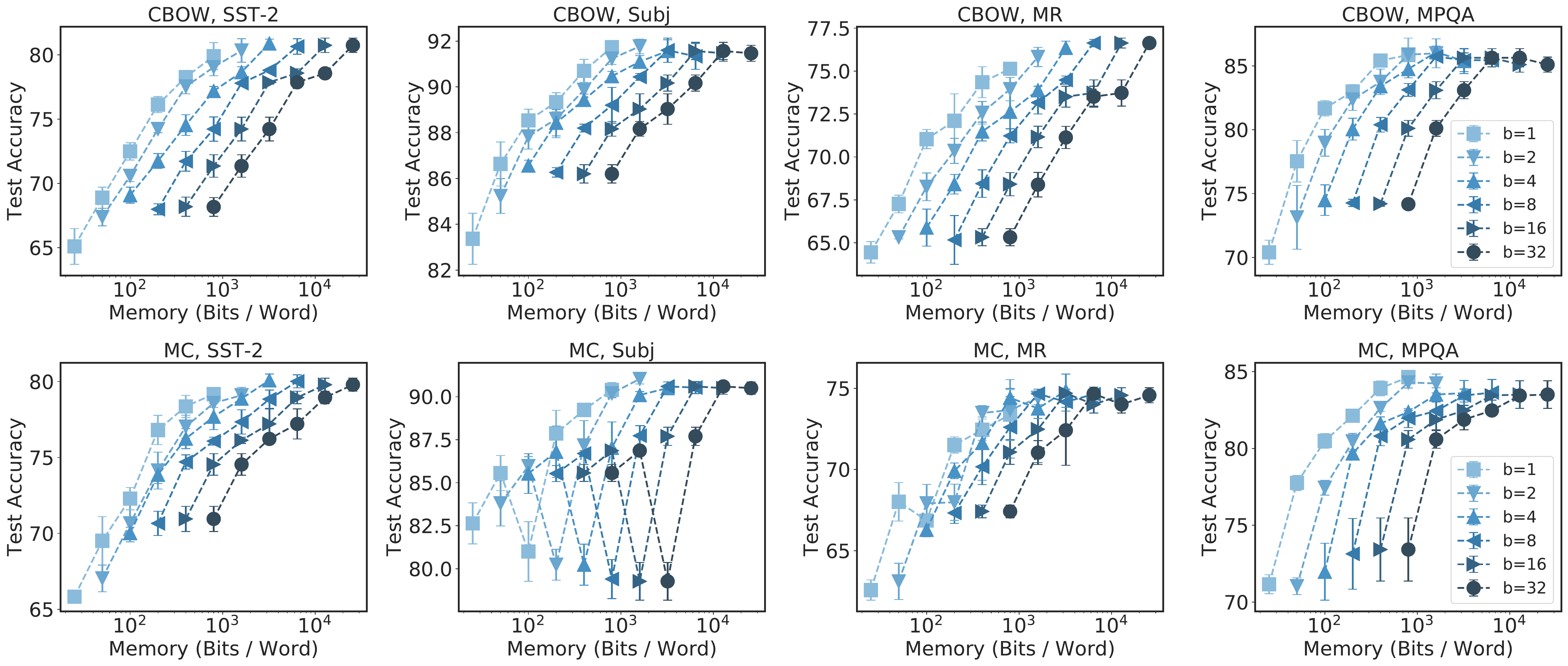}
    \caption{Quality-memory tradeoffs}
    \end{subfigure}

    \centering
    \begin{subfigure}{\textwidth}
    \centering
    \includegraphics[width=\textwidth]{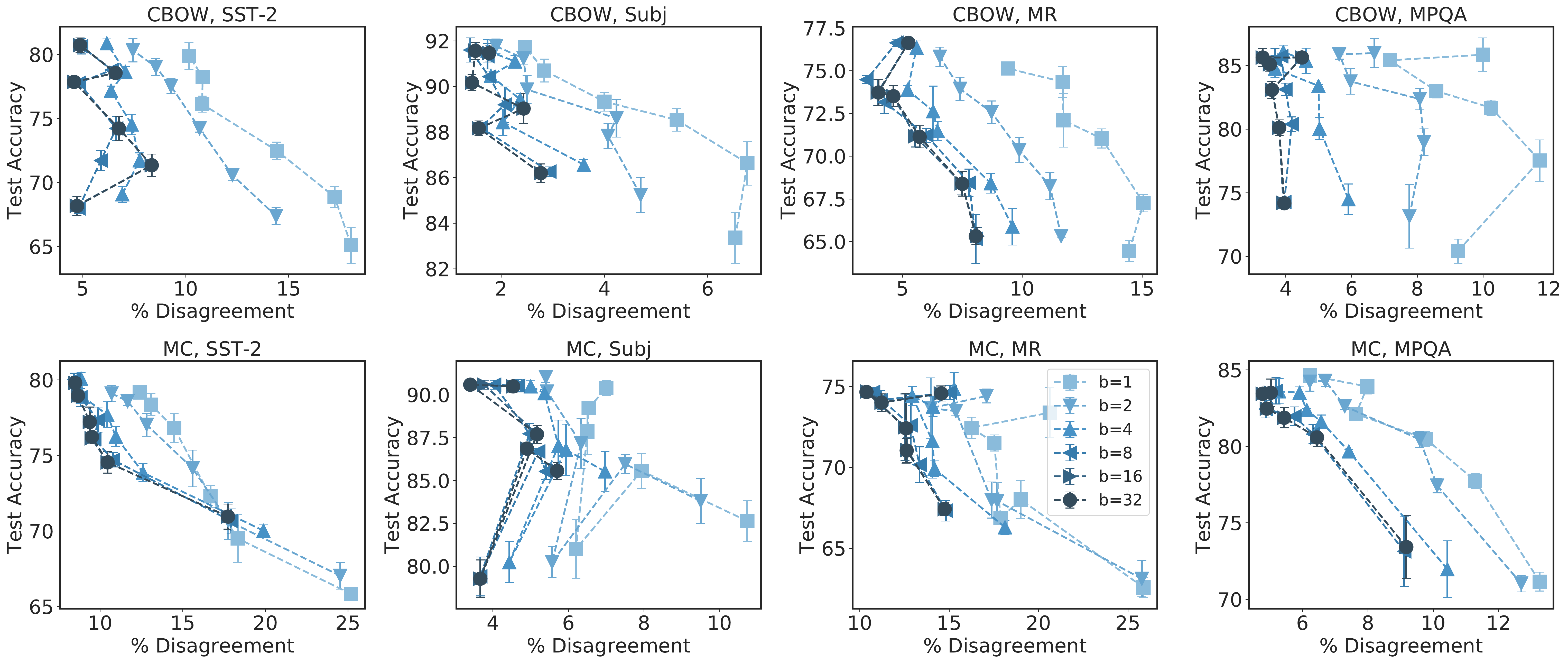}
    \caption{Quality-stability tradeoffs}
    \end{subfigure}
\caption{Quality tradeoffs for the sentiment analysis tasks with CBOW (top) and MC (bottom) embeddings for varying dimension-precision combinations.}
\label{fig:sa-quality}
\end{figure*}

\begin{figure}[!h]
    \centering
    \begin{subfigure}{0.45\textwidth}
    \centering
        \includegraphics[width=\textwidth]{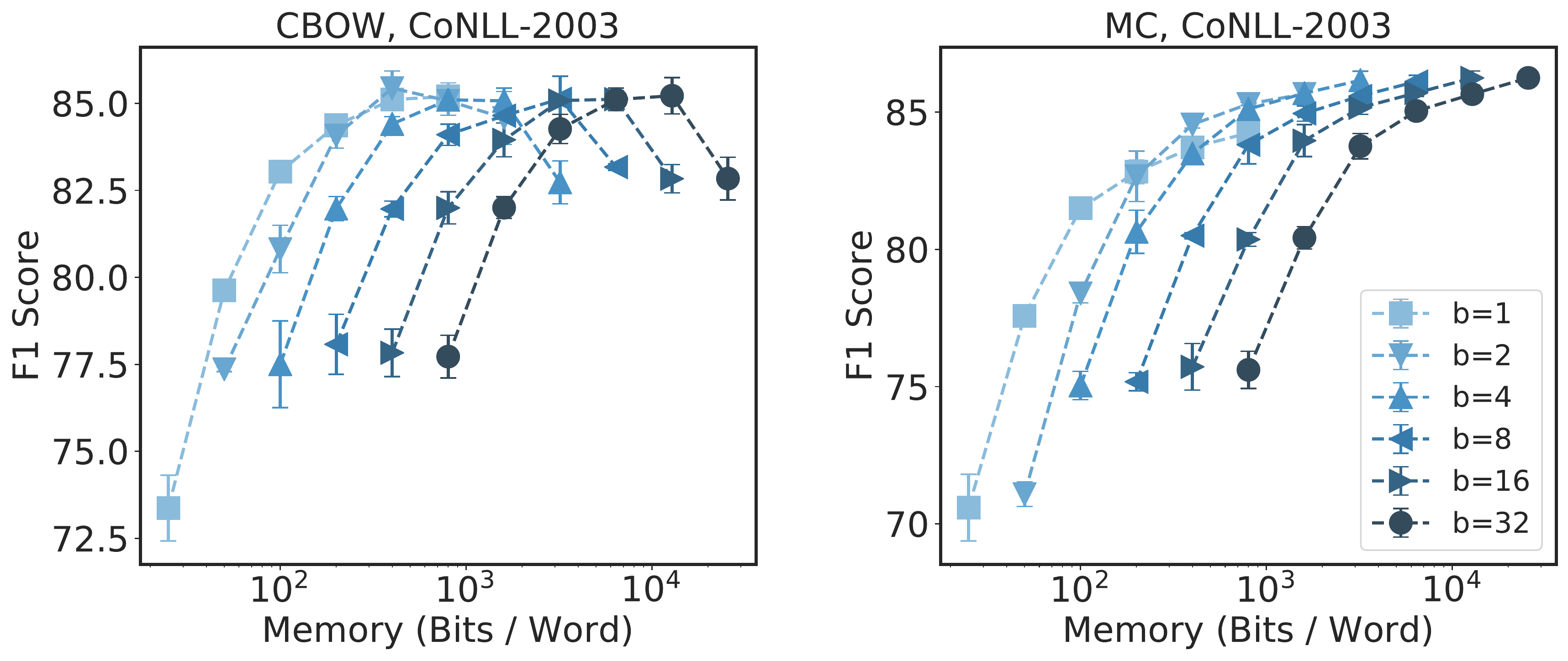}
    \caption{Quality-memory tradeoffs}
    \end{subfigure}
    \centering
    \hspace{.05\textwidth}
    \begin{subfigure}{0.45\textwidth}
    \centering
    \includegraphics[width=\textwidth]{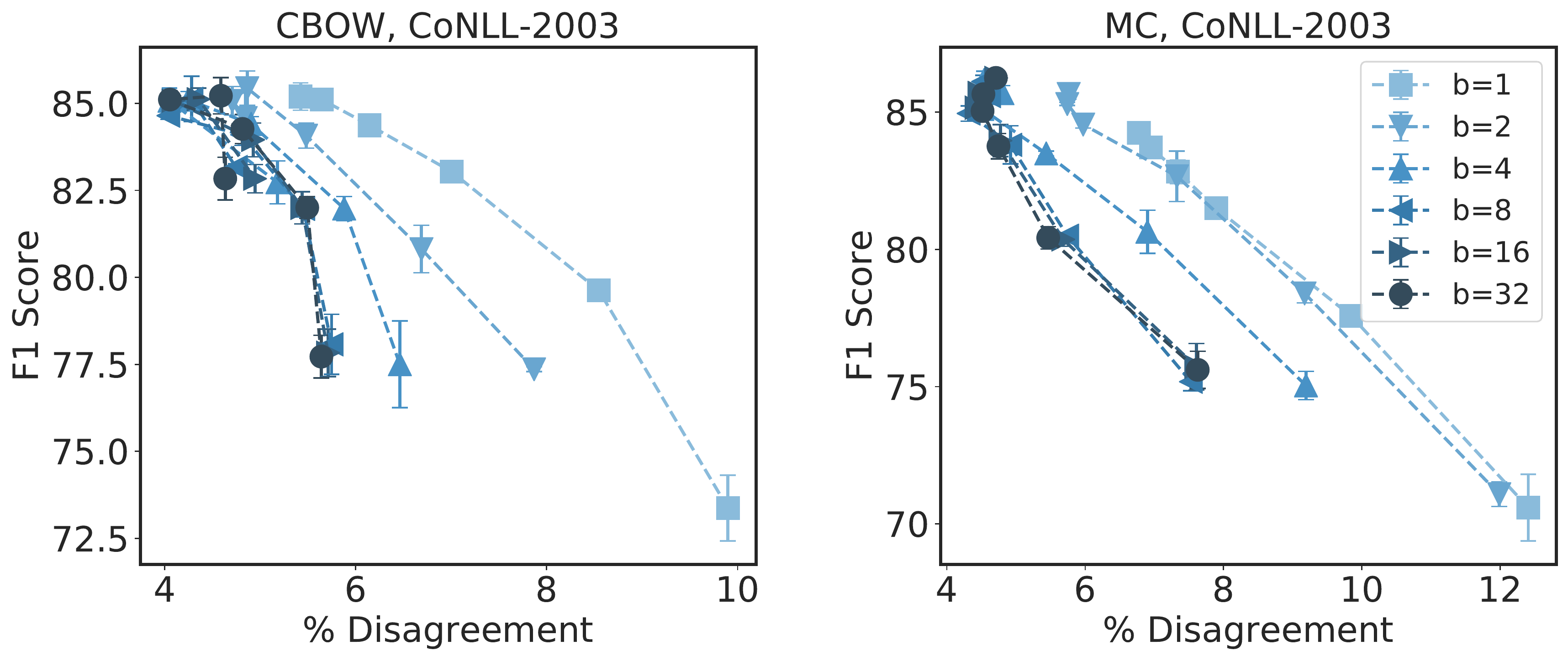}
    \caption{Quality-stability tradeoffs}
    \end{subfigure}
\caption{Quality tradeoffs for the NER task with CBOW and MC embeddings for varying dimension-precision
combinations.}
\label{fig:ner-quality}
\end{figure}

We also evaluate the quality-memory tradeoffs and quality-stability tradeoffs for CBOW and MC embedding algorithms, finding that like stability, the quality also increases with the embedding memory. In Figures~\ref{fig:sa-quality}~(a) and \ref{fig:ner-quality}~(a), we show the quality-memory tradeoff across sentiment analysis and NER tasks and CBOW and MC embedding algorithms for different dimension-precision combinations. We see that the dimension tends to impact the quality significantly more than the precision (i.e., the change in dimension for a fixed precision affects the quality more than the change in precision for a fixed dimension affects the quality). Recall that in contrast, for instability, we saw that the precision actually had a slightly greater effect than the dimension in Section~\ref{sec:joint}. In Figures~\ref{fig:sa-quality}~(b) and \ref{fig:ner-quality}~(b) we also show the quality-stability tradeoffs. For many of the sentiment analysis tasks, there is not significant evidence of a strong relationship between the two; however, for the NER task, we can clearly see that as the instability increases, the quality decreases. For several of the tasks (e.g., CBOW, MR; CBOW, MPQA), we can see that for different precisions (i.e., lines), the instability changes significantly, but the quality is relatively constant. This aligns with the previous observation that the precision tends to impact the instability more than it does the quality. In a similar way, for different dimensions (i.e., points), we see that the quality can change significantly while the instability may stay relatively constant, especially for higher precisions (e.g., CBOW, SST-2, CBOW, MPQA).

\subsection{Selecting Hyperparameters for Embedding Distance Measures}
\label{sec:measure-hyperparams}
The eigenspace instability measure and the measure each have a single hyperparameter to tune. For the eigenspace instability measure, $\alpha$ determines how important the directions of the eigenvalues of high variance are. For the measure, $k$ determines how many neighbors are compared for each query word. To tune these hyperparameters, we compute the Spearman correlation between the embedding distance measure and the downstream prediction disagreement on the validation datasets for the five tasks in our study and MC and CBOW embedding algorithms. In Table~\ref{tab:alpha-vals} we report the average Spearman correlation for different values of $\alpha$ for the eigenspace instability measure where we see $\alpha=3$ is the top-performing value. In Table~\ref{tab:k-vals} we report the average Spearman correlation for different values of $k$ for the k-NN measure, where we see $k=5$ is the top-performing value. Based on these results, we use $\alpha=3$ and $k=5$ for our experiments throughout the paper.
\vspace{-4mm}
\begin{table}[!htb]
    \caption{Average Spearman correlation $\rho$ values for different values of $\alpha$ for the eigenspace instability measure (a) and $k$ for the k-NN measure (b). Top value bolded.}
    \medskip
    \small
    \begin{subtable}{.5\linewidth}
        \centering
        \caption{$\alpha$ for the eigenspace instability measure}
        \medskip
        \small
        \label{tab:alpha-vals}
        \begin{tabular}{lc}
        \toprule
        $\alpha$ & $\rho$ \\
        \midrule
        0        & -0.350 \\
        1        & -0.067 \\
        2        & 0.498  \\
        3        & \textbf{0.751}  \\
        4        & 0.748  \\
        5        & 0.741  \\
        6        & 0.738  \\
        7        & 0.739  \\
        8        & 0.739 \\
        \bottomrule
        \end{tabular}
    \end{subtable}%
    \begin{subtable}{.5\linewidth}
        \centering
        \caption{$k$ for the k-NN measure}
        \medskip
        \small
        \label{tab:k-vals}
        \begin{tabular}{lc}
        \toprule
        $k$    & $\rho$ \\
        \midrule
        1    & 0.766 \\
        2    & 0.777 \\
        5    & \textbf{0.785} \\
        10   & 0.782 \\
        50   & 0.774 \\
        100  & 0.763 \\
        500  & 0.703 \\
        1000 & 0.675 \\
        \bottomrule
        \end{tabular}
    \end{subtable}
\end{table}
\vspace{-5mm}
\subsection{Predictive Performance of the Eigenspace Instability Measure}
\label{sec:app-corr}
\begin{figure*}[!h]
        \centering
        \begin{subfigure}{0.8\textwidth}
        \includegraphics[width=\textwidth]{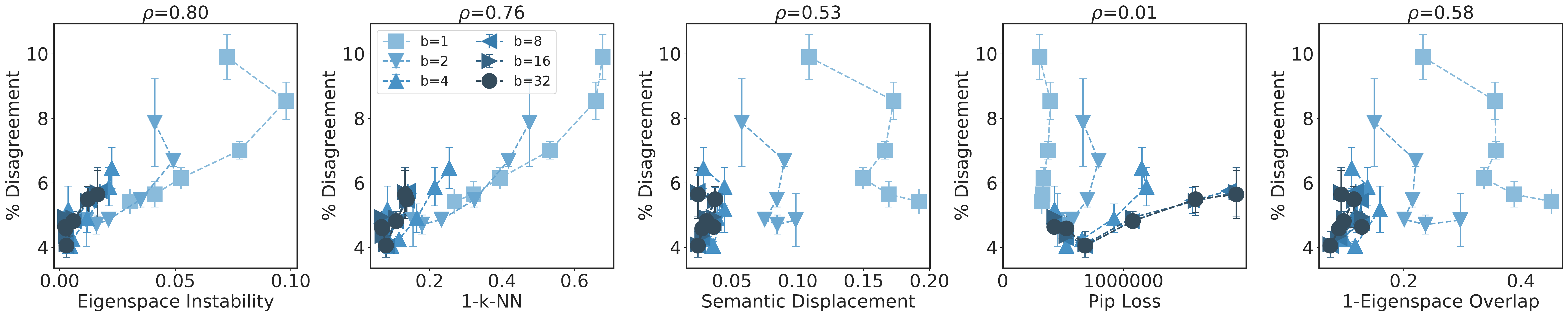}
        \caption{CBOW}
        \end{subfigure}
        \begin{subfigure}{0.8\textwidth}
            \includegraphics[width=\textwidth]{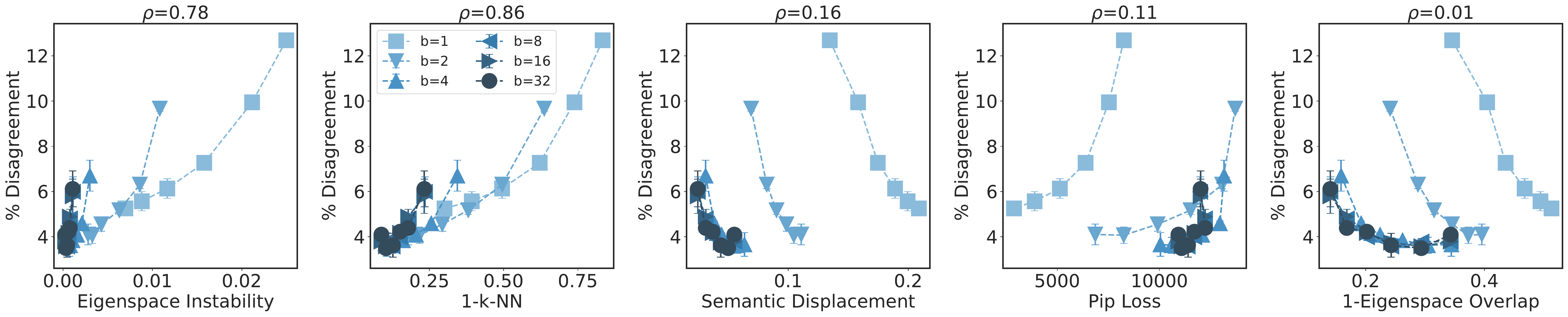}
            \caption{GloVe}
            \end{subfigure}
        \begin{subfigure}{0.8\textwidth}
        \includegraphics[width=\textwidth]{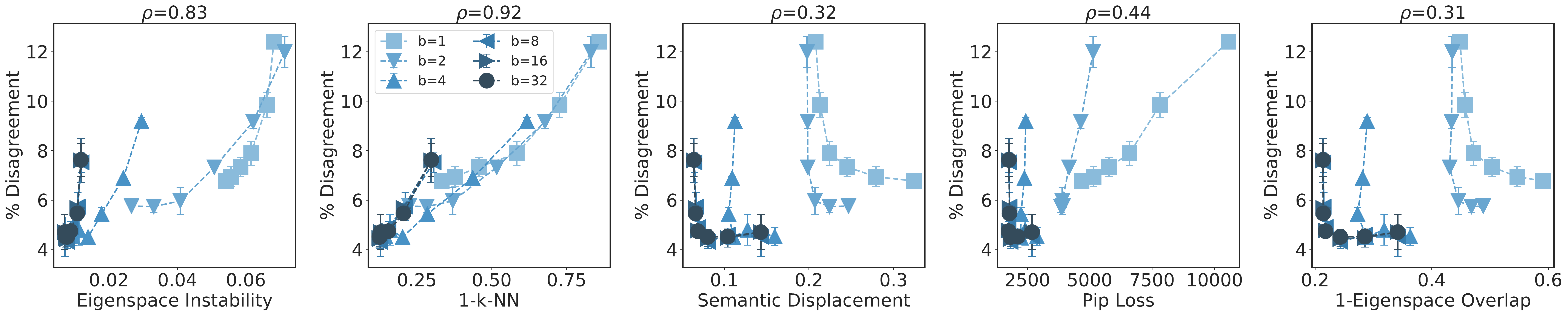}
        \caption{Matrix completion (MC)}
        \end{subfigure}
        \caption{Downstream instability versus embedding distance measures for the NER task on the CoNLL-2003 dataset. $\rho$ is the Spearman correlation between the embedding distance measure and the downstream instability. }
        \label{fig:corr-ner}
        \end{figure*}

We now provide additional results validating the strong relationship between the eigenspace instability measure and downstream instability. In addition to the Spearman correlation results we provide in Table~\ref{tab:corr-scores}, we visualize the downstream instability v. embedding distance measure results for the CoNLL-2003 NER task in Figure~\ref{fig:corr-ner} with CBOW and MC embeddings, taking the average over three seeds. We see that k-NN measure and the eigenspace instability measure achieve strong correlations since the lines are generally monotonically increasing for both CBOW and MC embedding algorithms.

\subsection{Embedding Distance Measures for Dimension-Precision Selection}

We first include Spearman correlation results and selection task results for CBOW, GloVe, and MC on the two additional downstream tasks--MR and MPQA--in Tables~\ref{tab:corr-scores-extra},~\ref{tab:sel-scores-extra}, and~\ref{tab:avg-case-extra}, where we see that the eigenspace instability measure and the k-NN measure continue to outperform the other measures.

\begin{table*}[h]
    \caption{Extended results for Section~\ref{sec:experiments} on MR and MPQA downstream tasks. Top-performing values are bolded.}
    \centering
    \vspace{3mm}
    \begin{subtable}{\linewidth}
    \caption{Spearman correlation scores between embedding distance measures and downstream prediction disagreement across varying dimension-precision pairs for the embedding. Downstream models are trained for sentiment analysis (MR, MPQA) tasks.}
    \medskip
    \small
    \label{tab:corr-scores-extra}
    \begin{tabularx}{\textwidth}{lYYYYYY}
    \toprule
    \textit{Downstream Task}  & \multicolumn{3}{c}{MR}  & \multicolumn{3}{c}{MPQA} \\
    \midrule
    \textit{Embedding Algorithm} & CBOW  & GloVe & MC  & CBOW & GloVe & MC   \\
    \cmidrule(lr){0-0}
    \cmidrule(lr){2-4}
    \cmidrule(lr){5-7}
    Eigenspace Instability   & \bf{0.86}	& 0.77	& 0.72	& 0.75	& 0.84	 & 0.85 \\
    $1 - \text{k-NN}$                     & 0.85	& \bf{0.86}	& \bf{0.74}	& \bf{0.77}	& \bf{0.92}	 & \bf{0.94} \\
    Semantic Displacement    & 0.63	& 0.10	& 0.59	& 0.68	& 0.10	 & 0.29 \\
    PIP Loss                 & -0.11	& 0.14	& 0.66	& -0.38	& 0.16	 & 0.42 \\
    $1-\text{Eigenspace Overlap}$     & 0.66	& -0.06	& 0.56	& 0.68	& -0.07	 & 0.27 \\
    \bottomrule
    \end{tabularx}
\end{subtable}
\vspace{3mm}

    \begin{subtable}{\linewidth}
    \caption{Selection error when using embedding distance measures to predict the most stable embedding dimension-precision parameters on downstream tasks. Downstream models are trained for sentiment analysis (MR, MPQA) tasks. }
    \medskip
    \small
    \label{tab:sel-scores-extra}
    \begin{tabularx}{\textwidth}{lYYYYYY}
    \toprule
    \textit{Downstream Task}  & \multicolumn{3}{c}{MR}  & \multicolumn{3}{c}{MPQA} \\
    \midrule
    \textit{Embedding Algorithm}                          & CBOW  & GloVe & MC  & CBOW & GloVe & MC  \\
    \cmidrule(lr){0-0}
    \cmidrule(lr){2-4}
    \cmidrule(lr){5-7}
    Eigenspace Instability    & \bf{0.14} & 0.20 & 0.24 & 0.22 & 0.18 & 	0.17 \\
    $1 - \text{k-NN}$                      & 0.15 & \bf{0.15} & \bf{0.22} & \bf{0.21} & \bf{0.12} & 	\bf{0.11} \\
    Semantic Displacement     & 0.26 & 0.48 & 0.29 & 0.24 & 0.49 & 	0.40 \\
    PIP Loss                  & 0.52 & 0.43 & 0.25 & 0.64 & 0.42 & 	0.33 \\
    $1-\text{Eigenspace Overlap}$      & 0.27 & 0.53 & 0.29 & 0.25 & 0.55 & 	0.41 \\
    \bottomrule
    \end{tabularx}
\end{subtable}
\begin{subtable}{\linewidth}
    \vspace{3mm}
    \caption{Average difference (absolute percentage) to the oracle downstream instability when using embedding distance measures as the selection criteria for dimension and precision parameters under fixed memory budgets.
    }
    \medskip
    \small
    \label{tab:avg-case-extra}
    \begin{tabularx}{\textwidth}{lYYYYYY}
        \toprule
        \textit{Downstream Task}  & \multicolumn{3}{c}{MR}  & \multicolumn{3}{c}{MPQA} \\
        \midrule
        \textit{Embedding Algorithm}                          & CBOW  & GloVe & MC  & CBOW & GloVe & MC  \\
        \cmidrule(lr){0-0}
        \cmidrule(lr){2-4}
        \cmidrule(lr){5-7}
    Eigenspace Instability   & \bf{0.69} & 1.35 & 1.03 & 0.34 & 0.87 & 0.44 \\
    $1 - \text{k-NN}$ & 0.71 & \bf{1.32} & \bf{0.57} & \bf{0.32} & \bf{0.86} & \bf{0.33} \\
    Semantic Displacement & 0.72 & 2.72 & 2.02 & 0.56 & 2.43 & 1.33 \\
    PIP Loss & 3.33 & 2.03 & 1.96 & 3.24 & 1.40 & 1.05 \\
    $1-\text{Eigenspace Overlap}$ & 0.84 & 2.72 & 2.02 & 0.51 & 2.43 & 1.25 \\
    High Precision & 1.55 & 2.72 & 2.07 & 0.40 & 2.43 & 1.49 \\
    Low Precision   & 3.33 & 2.03 & 4.09 & 3.24 & 1.40 & 0.66 \\
    \bottomrule
    \end{tabularx}
\end{subtable}
    \end{table*}

\label{sec:wc-robustness}
We also evaluate the the worst-case performance of the embedding distance measures when used as a selection criterion for dimension-precision parameters.  First, on the easier task of choosing the more stable dimension-precision pair out of two choices, we define the worst-case performance as the maximum increase in instability that may occur by using the embedding distance measure to choose the dimension-precision parameters (rather than the ground truth choice). On the more challenging task of choosing the most stable dimension-precision pair under a memory budget, we define the worst-case performance as the worst-case absolute percentage error to the oracle parameters under a given memory budget. We see in Tables~\ref{tab:sel-scores-robustness} and \ref{tab:worst-case} that the eigenspace instability measure and k-NN measure are the top-performing measures overall across both tasks.

\begin{table*}[h]
    \caption{Worst-case absolute percentage error when using each embedding distance measure to predict the most stable embedding parameters on downstream tasks over of all pairs of parameters. Downstream models are trained for sentiment (SST-2, Subj) and NER (CoNLL-2003) tasks. Lowest errors are bolded.}
    \medskip
    \small
    \label{tab:sel-scores-robustness}
    \begin{tabularx}{\textwidth}{lYYYYYYYYY}
        \toprule
        \textit{Downstream Task}  & \multicolumn{3}{c}{SST-2}  & \multicolumn{3}{c}{Subj} &  \multicolumn{3}{c}{CoNLL-2003} \\
        \midrule
        \textit{Embedding Algorithm}                          & CBOW  & GloVe & MC  & CBOW & GloVe & MC   & CBOW  & GloVe & MC  \\
        \cmidrule(lr){0-0}
        \cmidrule(lr){2-4}
        \cmidrule(lr){5-7}
        \cmidrule(lr){8-10}
    Eigenspace Instability & \bf{10.43}	& 6.48	& 13.18	& 3.50	& \bf{3.00}	& \bf{3.40}	 & 3.30	 & 4.04	& 4.11 \\
    $1 - \text{k-NN}$                 & \bf{10.43}	& \bf{4.78}	& \bf{11.75}	& \bf{2.80}	& \bf{3.00}	& \bf{3.40}	 & \bf{2.17}	 & \bf{2.39}	& \bf{3.16} \\
    Semantic Displacement  & 11.70	& 9.23	& 16.80	& 5.40	& 6.00	& 7.10	 & 5.13	 & 8.05	& 7.03 \\
    PIP Loss               & 16.14	& 14.61	& 15.76	& 6.40	& 9.10	& 4.40	 & 6.78	 & 9.69	& 5.77 \\
    $1-\text{Eigenspace Overlap}$   & 12.69	& 11.92	& 16.80	& 5.50	& 8.50	& 7.10	 & 5.86	 & 9.23	& 7.03 \\
    \bottomrule
    \end{tabularx}
\end{table*}
\vspace{-2mm}
\begin{table*}[h]
\caption{Worst-case absolute percentage error to the oracle downstream instability when using embedding distance measures as the selection criteria for dimension and precision parameters. Smallest errors are bolded.}
\medskip
\small
\label{tab:worst-case}
    \begin{tabularx}{\textwidth}{lYYYYYYYYY}
        \toprule
        \textit{Downstream Task}  & \multicolumn{3}{c}{SST-2}  & \multicolumn{3}{c}{Subj} &  \multicolumn{3}{c}{CoNLL-2003} \\
        \midrule
        \textit{Embedding Algorithm}                          & CBOW  & GloVe & MC  & CBOW & GloVe & MC   & CBOW  & GloVe & MC  \\
        \cmidrule(lr){0-0}
        \cmidrule(lr){2-4}
        \cmidrule(lr){5-7}
        \cmidrule(lr){8-10}
Eigenspace Instability & 3.08	& \bf{4.78}	& 11.37 & \bf{1.80} & \bf{2.50} & \bf{2.40} & 	\bf{0.84} & \bf{1.96} & 	1.73 \\
$1 - \text{k-NN}$                 & 3.02	& \bf{4.78}	& 11.37 & \bf{1.80} & \bf{2.50} & 	2.60 & 	1.29 & 	\bf{1.96} & 	\bf{1.01} \\
Semantic Displacement  & \bf{2.47}	& 5.82	& 13.95 & \bf{1.80} & 2.70 & 	3.10 & 	1.73 & 	3.07 & 	3.92 \\
PIP Loss               & 7.96	& 6.37	& 13.95 & 3.30 & 3.10 & 	3.10 & 	2.02 & 	2.00 & 	3.92 \\
$1-\text{Eigenspace Overlap}$   & 10.43	& 5.82	& 13.95 & \bf{1.80} & 2.70 & 	3.10 & 	1.29 & 	3.07 & 	3.92 \\
High Precision         & 10.43	& 5.82	& 13.95 & \bf{1.80} & 2.70 & 	3.10 & 	2.03 & 	3.07 & 	3.92 \\
Low Precision          & 7.96	& 6.37	& \bf{5.33} & 3.30 & 3.10 & 	4.60 & 	2.02 & 	2.00 & 	2.33 \\

\bottomrule
\end{tabularx}
\end{table*}

\subsection{Knowledge Graph Embeddings}
\label{sec:app-kg-exp}

\begin{figure}[!h]
    \centering
    \includegraphics[width=0.22\textwidth]{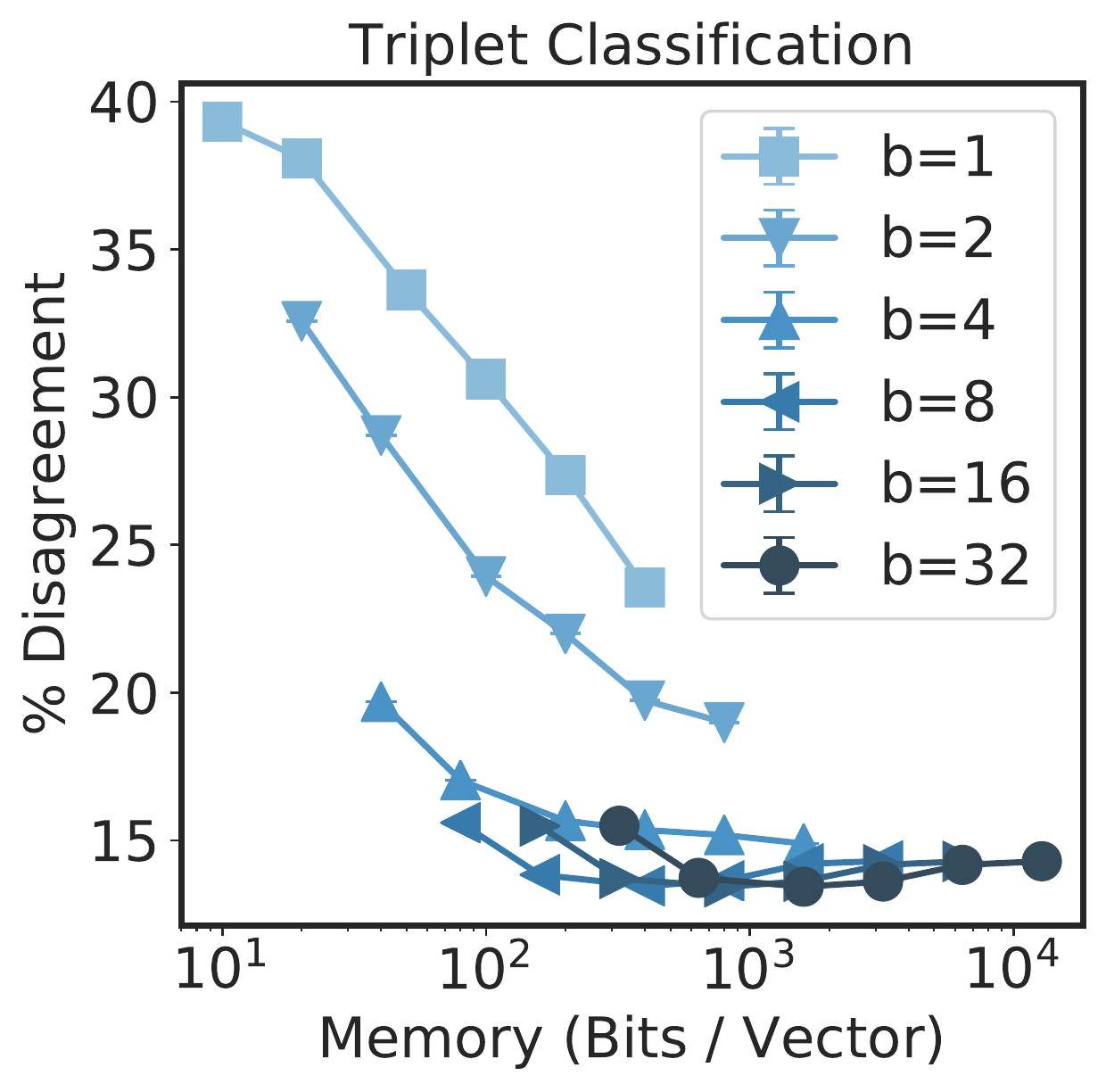}
\caption{Stability of triplet classification when evaluating embeddings when tuning the threshold for the task per dataset.}
\label{fig:kg-new-thresh}
\end{figure}

In Section~\ref{sec:kg}, we showed that as the memory of the TransE embedding increases, the instability on link prediction and triplet classification task decreases; we now experiment with a modified setup for the triplet classification experiments. In Figure~\ref{fig:kg-new-thresh}, we use thresholds tuned per dataset (in Figure~\ref{fig:transe} (right) we use the \emph{same threshold} on both the FB15K-95 and FB15K dataset) and see that the stability-memory tradeoffs are less pronounced for higher precisions.

\subsection{Contextual Word Embeddings}
\label{sec:app-cwe-exp}

\begin{figure*}[!h]
    \centering
    \begin{subfigure}{0.8\textwidth}
\centering
 \includegraphics[width=\textwidth]{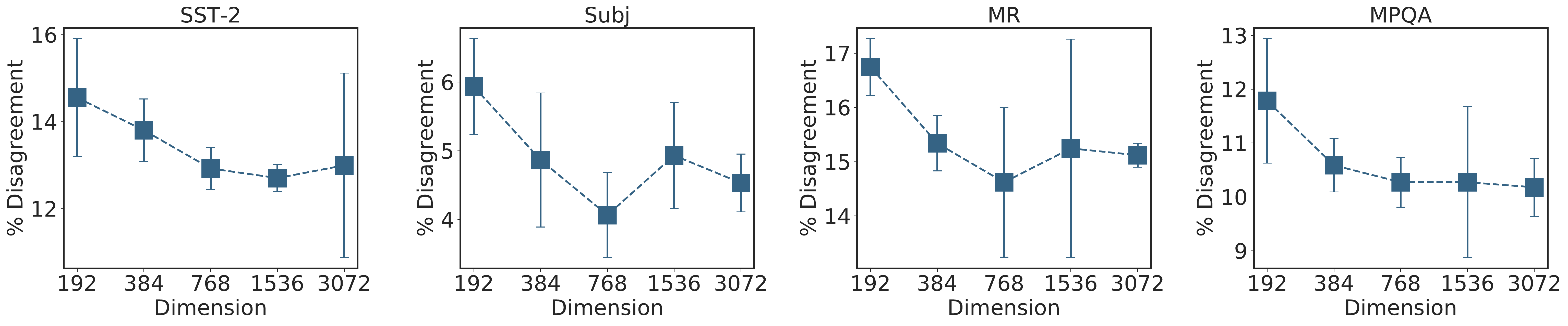}

 \caption{Effect of Dimension}
 \label{fig:bert_stab_dim}
 \end{subfigure}

\begin{subfigure}{0.8\textwidth}
\centering
\includegraphics[width=\textwidth]{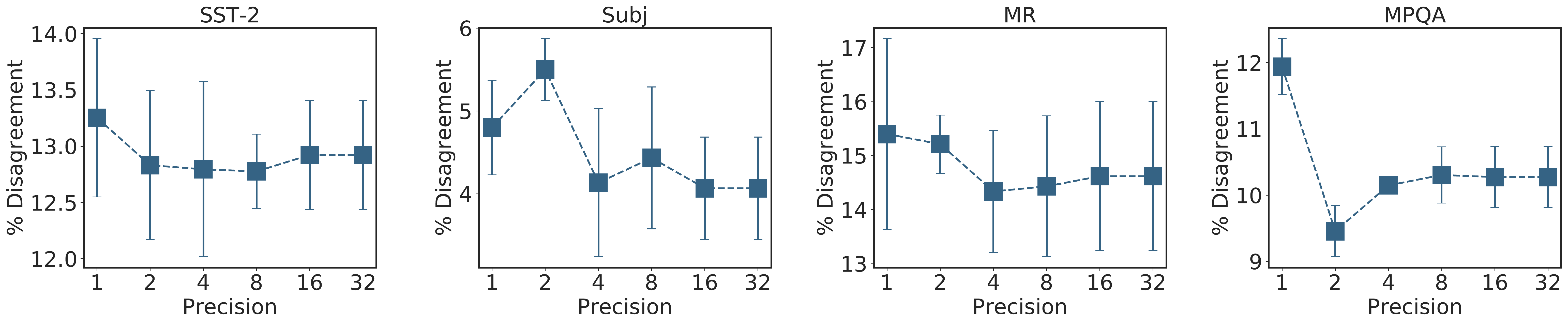}
 \caption{Effect of Precision}
 \label{fig:bert_stab_prec}
\end{subfigure}
\caption{Downstream instability of BERT embeddings on sentiment analysis with varying (a) output dimension, and (b) precision.\vspace{-3mm}}
\end{figure*}

We include the plots for the contextual word embedding experiments with BERT embeddings in Figures~\ref{fig:bert_stab_dim} and \ref{fig:bert_stab_prec} for dimension and precision, respectively. As discussed in Section~\ref{sec:bert}, although noisier than the trends with pre-trained word embeddings, we see that generally as the dimension and precision increase, the downstream instability decreases.

\label{app:bert}

%% file: limitations.tex
\section{Robustness of Trends}
\label{sec:robustness}
We explore the robustness of our study by providing preliminary investigation on the effect of subword embedding algorithms, more complex downstream models, other sources of randomness introduced by the downstream model (e.g., model initialization and sampling order), fine-tuning embeddings on downstream instability, and the downstream model learning rate.

\subsection{Subword Embeddings}
\begin{figure}[!h]
    \centering
        \includegraphics[width=0.48\textwidth]{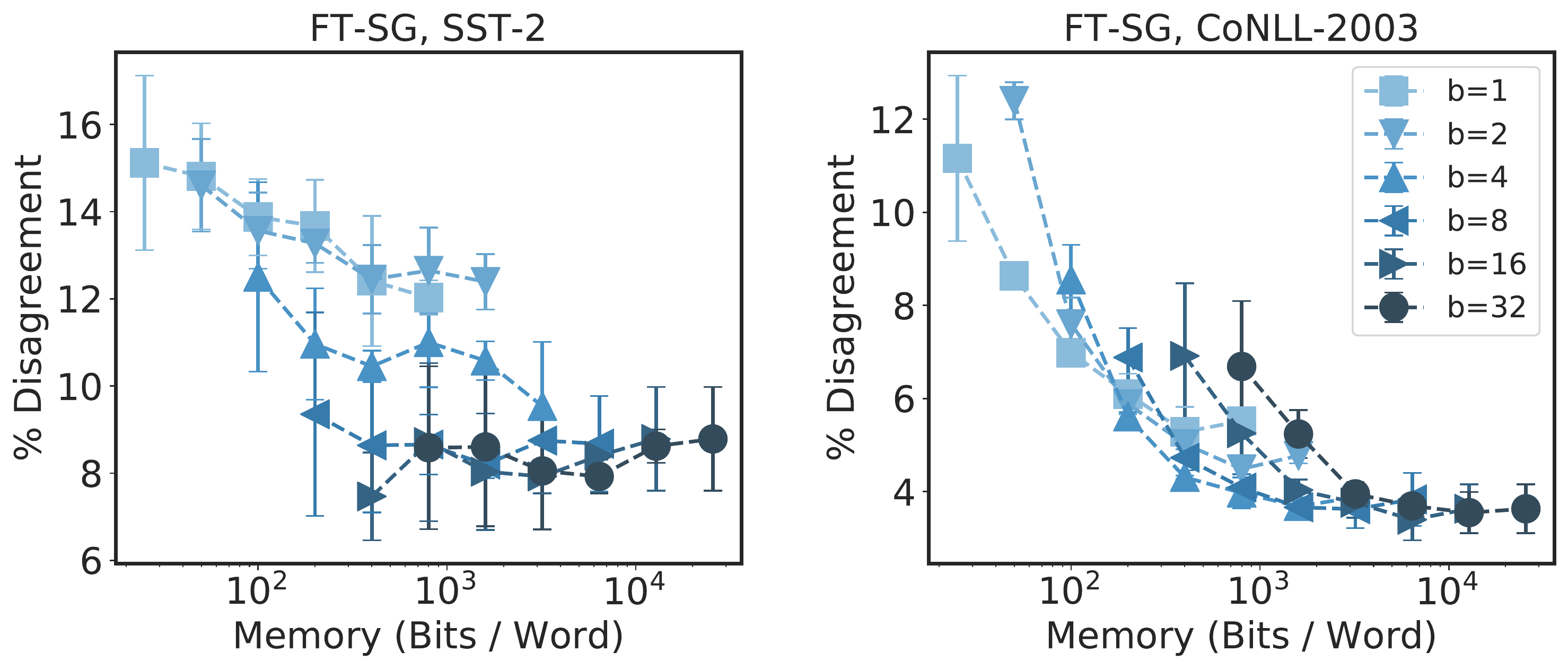}
    \caption{Downstream instability of fastText skipgram embeddings on SST-2 and CoNLL-2003 tasks.}
    \label{fig:ft-sg}
    \end{figure}

We experiment with fastText~\cite{bojanowski2016enriching} embeddings to evaluate if the stability-memory tradeoff holds on subword embedding methods. In Figure~\ref{fig:ft-sg}, we show that we can see that overall as the memory increases, the downstream instability decreases on the SST-2 and CoNLL-2003 tasks. However, the trend with respect to dimension is weaker on the SST-2 task at larger precisions.

\begin{figure}[!h]
    \centering
    \begin{subfigure}{0.4\textwidth}
    \centering
        \includegraphics[width=\textwidth]{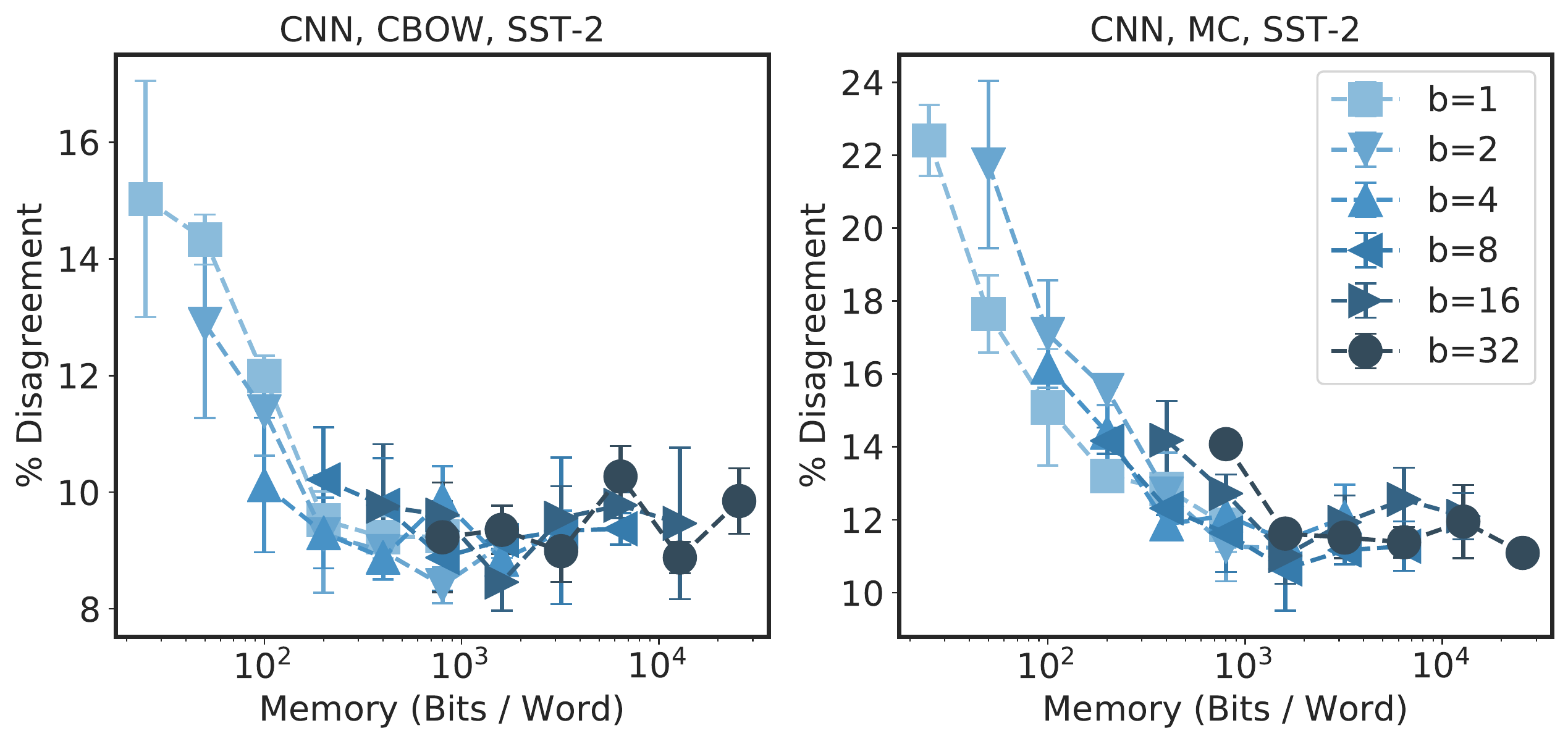}
    \caption{SST-2 task with a CNN.}
    \label{fig:cnn-cbow-mc}
    \end{subfigure}
    \hspace{.05\textwidth}
    \begin{subfigure}{0.4\textwidth}
    \centering
        \includegraphics[width=\textwidth]{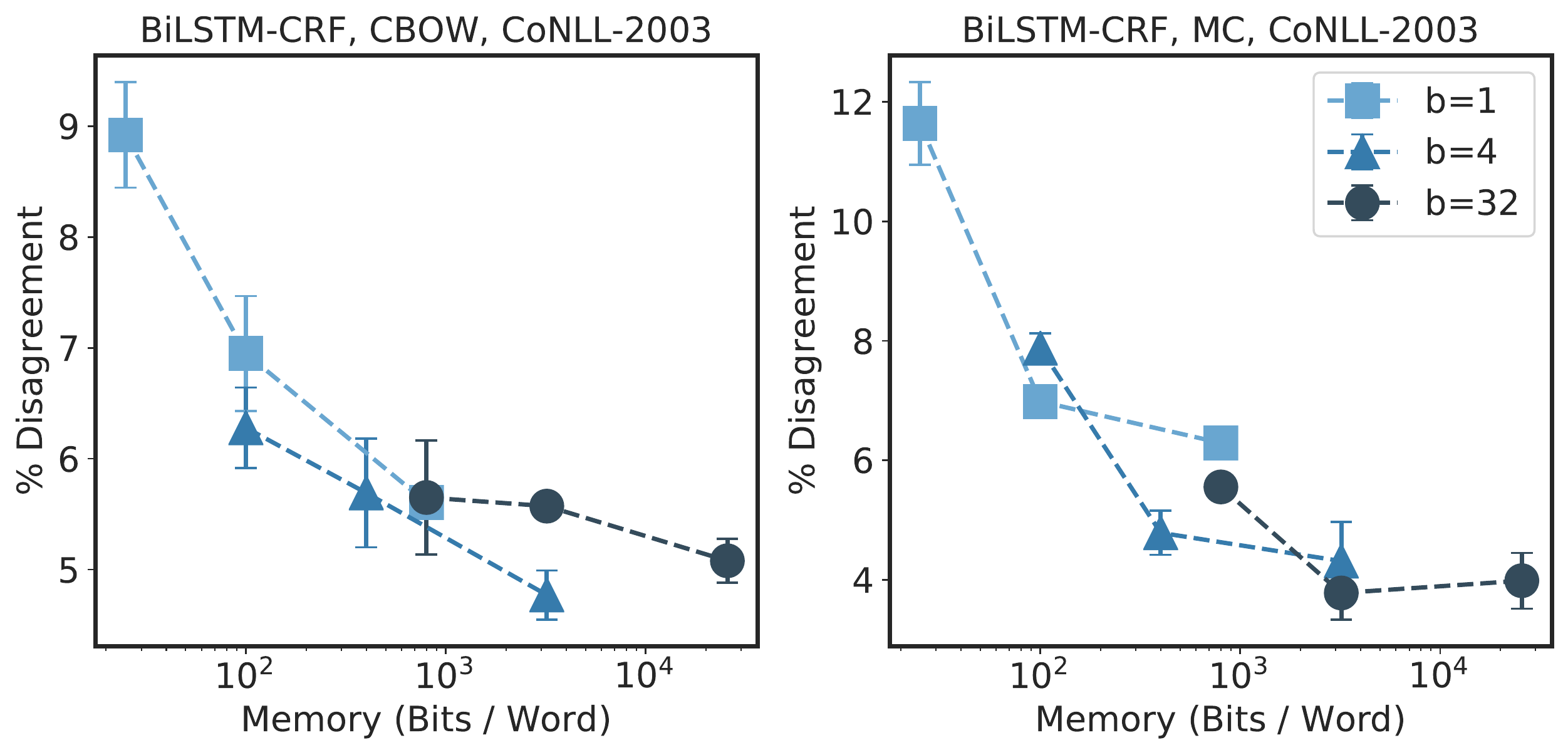}
    \caption{NER task with a BiLSTM-CRF.}
    \label{fig:crf}
    \end{subfigure}
    \caption{Downstream instability of CBOW and MC embeddings on more complex downstream models.}
\end{figure}

\subsection{Complex Downstream Models}
\label{sec:complex-ds}
In the main text, our primary downstream models are a simple linear bag-of-words model for sentiment analysis and a single layer BiLSTM for NER. We now demonstrate that complex downstream models such as CNNs or BiLSTM-CRFs can still demonstrate the stability-memory tradeoffs, such that as the memory increases, the instability decreases. In Figure~\ref{fig:cnn-cbow-mc}, we show that when using a CNN for the SST-2 sentiment analysis task, embeddings with very low memory budgets result in high instability. The instability quickly plateaus for memory budgets greater than $10^2$ for the CBOW embeddings, but continues to decrease until a memory budget of $10^3$ for the MC embeddings. The CNN architecture has one convolutional layer, with kernels of widths 3, 4, and 5, and 100 output channels. The convolutional layer is followed by a ReLU layer, a max-pooling layer, and finally a linear classification layer. We sweep the learning rate in a grid of \{1e-5, 0.0001, 0.001, 0.01, 0.1\} and choose the best learning rate by validation accuracy. Selected learning rates are shown in Table~\ref{tab:cnn-lr-params} and shared hyperparameters are shown in Table~\ref{tab:cnn-params}.

\begin{table}[!h]
    \centering
    \caption{(a) shows selected learning rates for the CNN architecture for sentiment analysis task on the SST-2 dataset per embedding algorithm. (b) shows training hyperparameters for the CNN architecture for sentiment analysis.}
    \vspace{2mm}
    \begin{subtable}{0.3\linewidth}
    \centering
        \caption{Tuned learning rates.}
        \medskip
        \small
        \label{tab:cnn-lr-params}
        \begin{tabular}{cc}
        \toprule
            CBOW       & MC \\
        \midrule
            0.0001  & 0.001   \\
        \bottomrule
        \end{tabular}
    \end{subtable}
    \begin{subtable}{0.6\linewidth}
    \caption{Shared training hyperparameters for the CNN.}
    \centering
    \medskip
    \small
    \label{tab:cnn-params}
    \begin{tabular}{lr}
    \toprule
        Hyperparameter       & Value \\
    \midrule
        Optimizer            & Adam   \\
        Batch size           & 32    \\
        Training epochs      & 100   \\
        Dropout             & 0.5   \\
    \bottomrule
    \end{tabular}
    \end{subtable}
    \end{table}

We now demonstrate that the BiLSTM-CRF also is subject to the stability-memory tradeoff, where as the dimension and precision increase, the instability decreases (Figure~\ref{fig:crf}). We use the same hyperparameters as in Table~\ref{tab:ner-shared-params} and repeat our setup for the BiLSTM with the CRF turned on for the CoNLL-2003 NER task. Due to the CRF being computationally expensive, we train a representative subset of points (dimensions in \{25, 100, 800\} and precisions in \{1, 4, 32\}). For each embedding algorithm, we grid search the learning rate for the BiLSTM-CRF with 400-dimensional embeddings in \{0.001, 0.01, 0.1, 1.0, 10.0\} and find that a learning rate of 0.1 is best for both embedding algorithms.

\subsection{Sources of Randomness Downstream}

We study the impact of two sources of randomness in the downstream model training---the model initialization seed and the sampling seed---on the downstream instability. First, we fix the embeddings and vary the model initialization seed and sampling seed independently. We vary the sampling order by shuffling the order of the batches in the training dataset. We compare the instability from these sources of randomness in the downstream model training to the instability from the embeddings. For each source of randomness downstream, we fix the embedding (using a single seed of the Wiki'17, full-precision, 400-dimensional embedding), and measure the instability between models trained with different random seeds. We repeat over three pairs of models and report the average. We see in Table~\ref{tab:random} that across four sentiment analysis tasks using the linear bag-of-words models, the sampling order seed introduces comparable instability to the change in embedding training data with full-precision, 400-dimensional embeddings, while the model initialization seed often contributes less instability. We note that using smaller memory budgets for the embeddings introduces much greater instability from the change in embedding training data, however, as shown in Figure~\ref{fig:ext-joint}.

In our experiments, we had also fixed the model initialization seeds and sampling order seeds to match that of the embedding, such that the seeds were the same between any two models we compared. We now remove this constraint, and vary the model initialization and sampling order seed of the model corresponding to the Wiki'18 embedding, such that no two models compared have the same seeds and otherwise repeat the experimental described in Section~\ref{sec:tradeoffs} and Appendix~\ref{sec:ds-setup}. In Figure~\ref{fig:seed-test-cbow-mc}, we see that the stability-memory tradeoffs continue to hold and the trends are very similar to when we fixed the seeds (Figure~\ref{fig:joint}). We note that many of the instability values themselves, particularly for CBOW, are slightly higher in Figure~\ref{fig:seed-test-cbow-mc} than they are in Figure~\ref{fig:joint}, likely due to the additional instability from the change in downstream model initialization and sampling seeds.

\begin{table}[!h]
    \centering
    \caption{Using fixed 400-dimensional Wiki'17 embeddings and a linear bag-of-words model for sentiment analysis, we vary the model initialization seed and sampling order seed to measure their effect on downstream instability, compared to changing the embedding training data. Values represent the average percentage disagreement between models, and the largest instability is bolded for each embedding algorithm and task combination.}
    \label{tab:random}
    \medskip
    \small
    \begin{tabular}{lcccccccc}
        \toprule
        \textit{Downstream Task}  &  \multicolumn{2}{c}{SST-2}  & \multicolumn{2}{c}{MR}  & \multicolumn{2}{c}{Subj} &  \multicolumn{2}{c}{MPQA}\\
   \midrule
        \textit{Embedding Algorithm}                          & CBOW  & MC   & CBOW  & MC   & CBOW  & MC   & CBOW  & MC    \\
    \cmidrule(lr){0-0}
\cmidrule(lr){2-3}
\cmidrule(lr){4-5}
\cmidrule(lr){6-7}
\cmidrule(lr){8-9}
Model Initialization Seed & 3.48      & 7.08      & 2.44      & 9.28       & 1.10      & 4.53      & 1.45      & 4.30 \\
Sampling Order Seed       & \bf{8.99} & 5.96      & \bf{5.87} & 10.09      & 0.57      & \bf{6.13} & \bf{5.59} & 1.92 \\
Embedding Training Data   & 6.59      & \bf{8.66} & 4.00      & \bf{11.22} & \bf{1.50} & 3.40      & 3.30      & \bf{4.78} \\
    \bottomrule
    \end{tabular}
    \end{table}

    \subsection{Effect of Fine-tuning Embeddings Downstream}

\begin{figure}[!h]
    \centering
    \begin{subfigure}{0.4\textwidth}
        \centering
        \includegraphics[width=\textwidth]{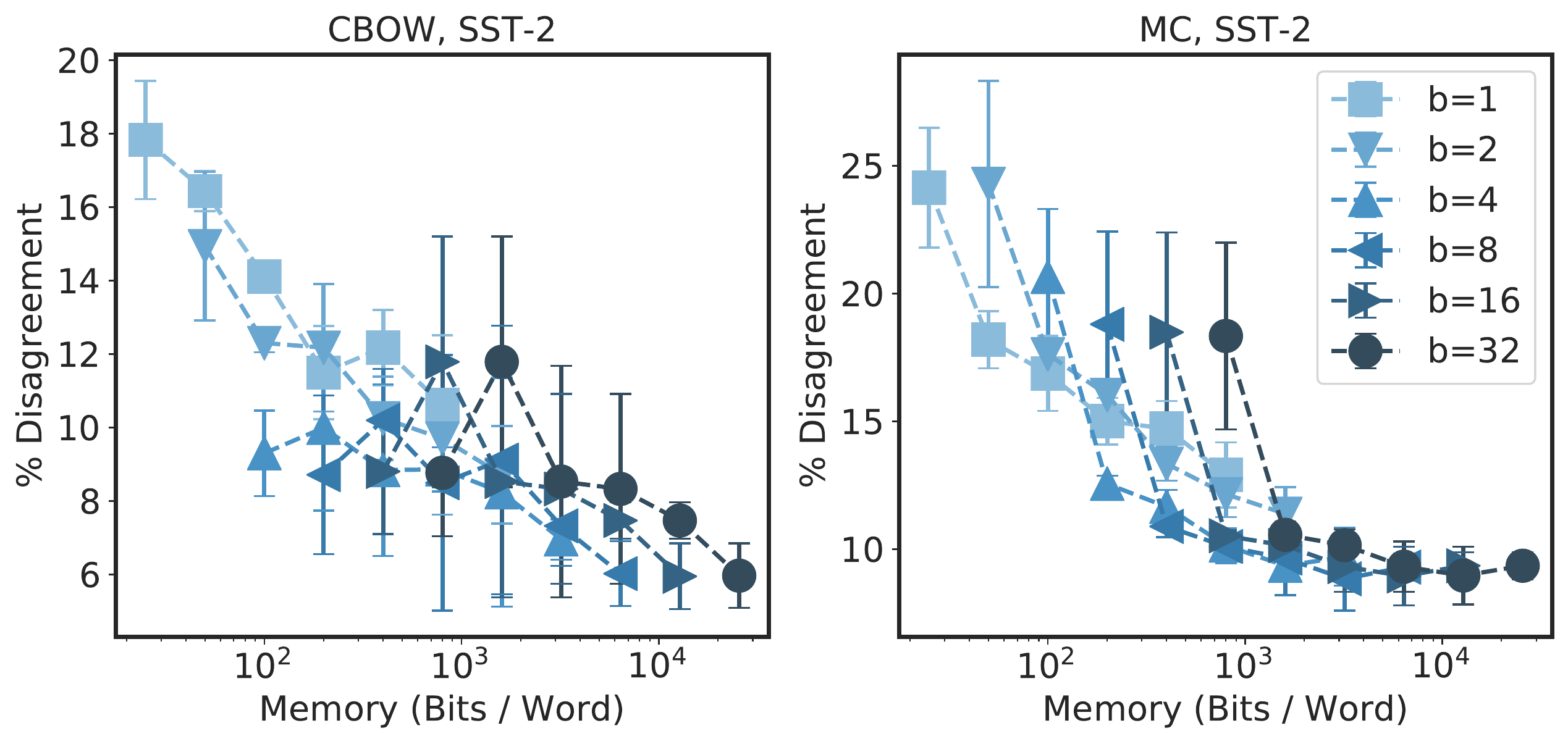}
    \caption{Relaxed seed constraint.}
    \label{fig:seed-test-cbow-mc}
    \end{subfigure}
    \hspace{.05\textwidth}
    \begin{subfigure}{0.4\textwidth}
    \centering
        \includegraphics[width=\textwidth]{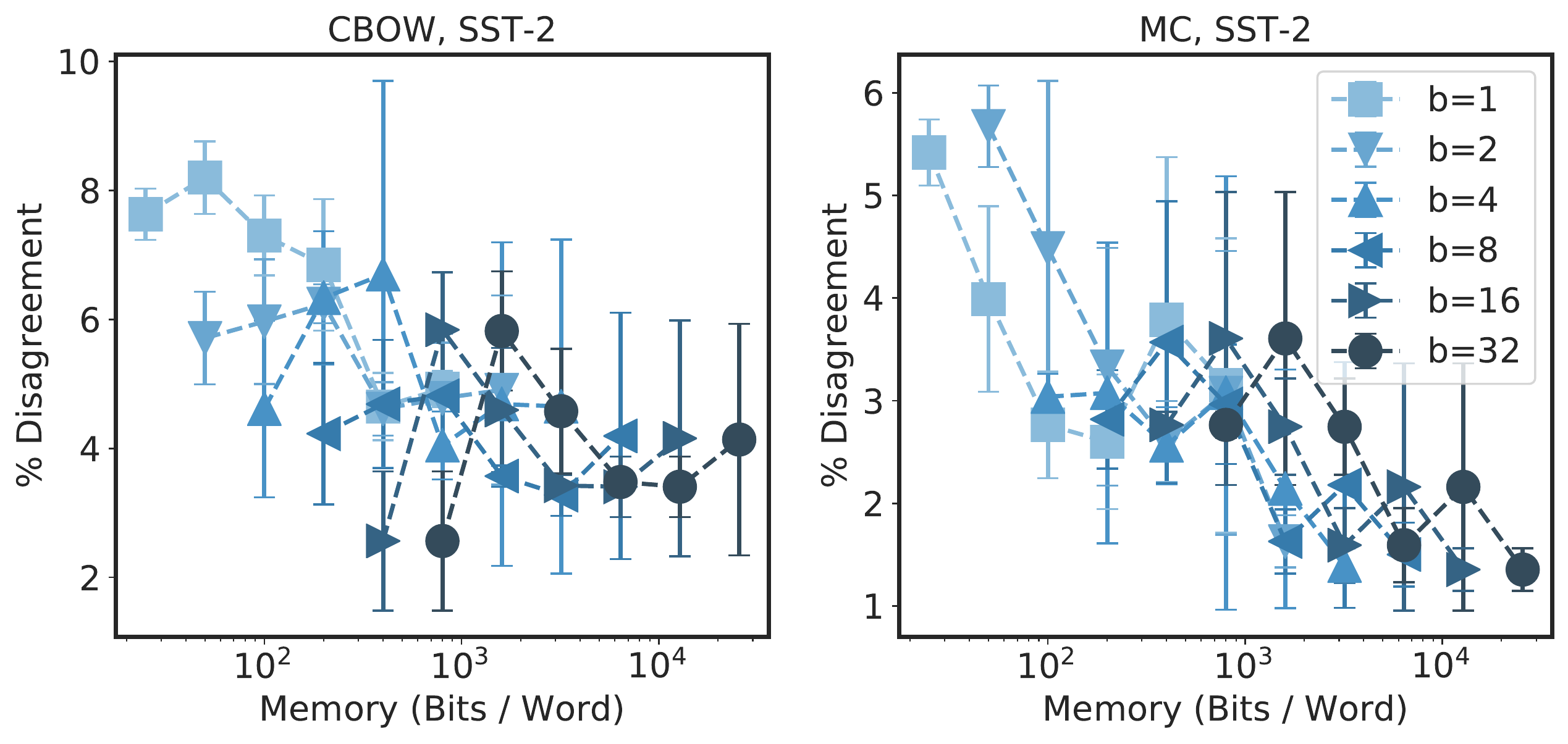}
    \caption{Fine-tuned embeddings.}
    \label{fig:finetune}
    \end{subfigure}
    \caption{Downstream instability of the SST-2 sentiment analysis task for different CBOW and MC embedding dimension-precision combinations when (a) we relax the constraint of having the model and sampling order seed be the same between models, and (b) embeddings are fine-tuned. The memory indicates the embedding memory \emph{prior to training the downstream model}, and embeddings are full-precision during downstream model training.}
\end{figure}

We study the impact of fine-tuning the embeddings downstream and find that the stability-memory tradeoff becomes noisier, but continues to hold under fine-tuning, and fine-tuning can dramatically help to decrease the downstream instability. In Figure~\ref{fig:finetune}, we show that as the memory increases, the instability generally decreases for both CBOW and MC embeddings, even when we allow the embeddings to be updated (i.e., fine-tuned) when training the downstream models. We note that we do not compress the embeddings during training in these experiments, therefore the memory denotes the memory required to store the embedding \emph{prior to training}. To perform the fine-tuning experiments, we follow the procedure described in Appendix~\ref{sec:ds-setup}, and perform an additional learning rate sweep per embedding algorithm with fine-tuning in the grid \{1e-5, 0.0001, 0.001, 0.01, 0.1, 1, 10\}. We found the optimal learning rate for both algorithms on the SST-2 sentiment analysis task with fine-tuning to be 0.0001. We also see that overall the instability decreases with fine-tuning compared to fixing the embeddings (as we did in Figure~\ref{fig:joint}). We note that the learning rate for the downstream model with MC and fine-tuning is smaller than with fixed embeddings, which may also contribute to the reduced instability; however, from Figure~\ref{fig:learning-rates}, the reduction in instability with fine-tuning still appears greater than that which can be achieved from a small change in learning rate alone.

\subsection{Effect of Downstream Learning Rate}
\label{sec:lr-effects}

\begin{figure}[!h]
    \centering
        \includegraphics[width=0.4\textwidth]{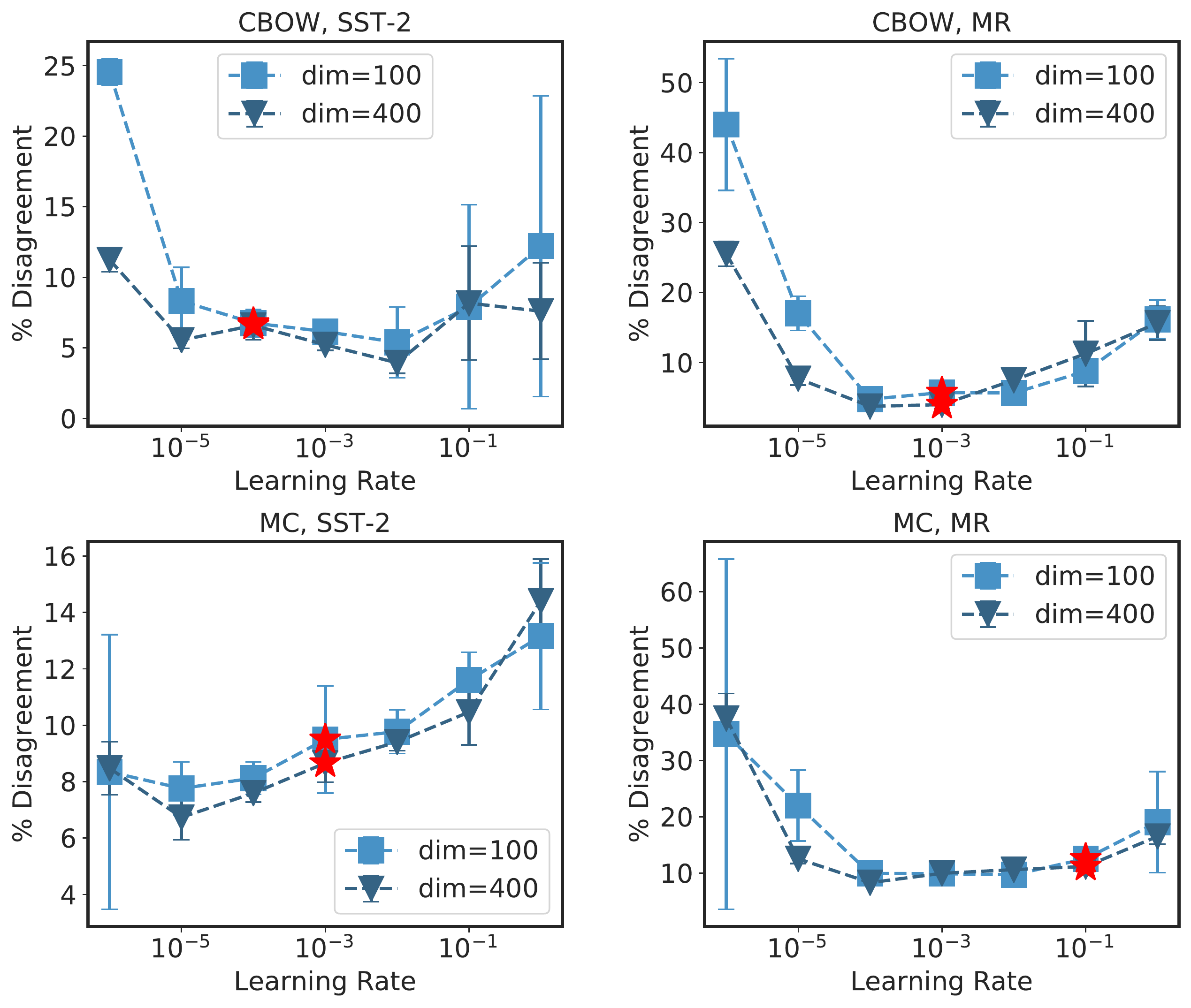}
    \caption{Downstream instability of CBOW and MC on the SST-2 and MR sentiment analysis tasks with various learning rates.}
    \label{fig:learning-rates}
\end{figure}

We now study the impact of the downstream model learning rate on the instability, showing that the learning rate of the downstream model is another factor that impacts the downstream instability.
In Figure~\ref{fig:learning-rates}, we show the instability of CBOW and MC embeddings on the SST-2 and MR sentiment analysis tasks when different learning rates are used for the downstream linear model. We mark the optimal learning rate by validation accuracy with a red star. We see that very small learning rates and very large learning rates tend to be the most unstable for both 100 and 400-dimensional embeddings. Moreover, the optimal learning rates do not significantly increase the instability compared to the other learning rates in our sweep. Since we see that the learning rate further contributes to the instability, we fix the learning rate over different precisions and dimensions in our main study to have a controlled setting to study the impact of dimension and precision on instability.

